\documentclass[conference]{IEEEtran}
\usepackage{times}

\usepackage[utf8]{inputenc}

\usepackage{graphicx}
\usepackage{soul}
\usepackage[dvipsnames,table,svgnames]{xcolor}

\usepackage[numbers]{natbib}
\usepackage{multicol}
\usepackage{multirow}
\usepackage[bookmarks=true]{hyperref}

\usepackage{caption} %
\usepackage{subcaption}
\definecolor{eggshell}{rgb}{0.94, 0.92, 0.84}
\definecolor{gray(x11gray)}{rgb}{0.75, 0.75, 0.75}
\usepackage{float}

\usepackage{array} %
\usepackage{booktabs} %
\usepackage[linesnumbered,ruled,lined, noend]{algorithm2e}

\usepackage{amssymb}%
\usepackage{amsmath}
\usepackage{pifont}%
\newcommand{\cmark}{\ding{51}}%

\usepackage[column=0]{cellspace} %
\begin{document}

\title{\textbf{IndustReal}: Transferring Contact-Rich Assembly Tasks from Simulation to Reality}

\author{\authorblockN{Bingjie Tang$^{\ast, 1}$, Michael A. Lin$^{\ast, 2}$, Iretiayo Akinola$^{3}$ , Ankur Handa$^{3}$, Gaurav S. Sukhatme$^1$, \\ Fabio Ramos$^{3, 4}$,  Dieter Fox$^{3, 5}$, Yashraj Narang$^{3}$ }
\authorblockA{$^{\ast}$Equal Contribution\\ 
$^1$University of Southern California, $^2$Stanford University, $^3$NVIDIA Corporation, \\ $^4$University of Sydney, $^5$University of Washington \\
}
}

\maketitle

\begin{abstract}

Robotic assembly is a longstanding challenge, requiring contact-rich interaction and high precision and accuracy. 
Many applications also require adaptivity to diverse parts, poses, and environments, as well as low cycle times. 
In other areas of robotics, simulation is a powerful tool to develop algorithms, generate datasets, and train agents. 
However, simulation has had a more limited impact on assembly.
We present IndustReal, a set of algorithms, systems, and tools that solve assembly tasks in simulation with reinforcement learning (RL) and successfully achieve policy transfer to the real world. Specifically, we propose 1) simulation-aware policy updates, 2) signed-distance-field rewards, and 3) sampling-based curricula for robotic RL agents.
We use these algorithms to enable robots to solve contact-rich pick, place, and insertion tasks in simulation.
We then propose 4) a policy-level action integrator to minimize error at policy deployment time.
We build and demonstrate a real-world robotic assembly system that uses the trained policies and action integrator to achieve repeatable performance in the real world.
Finally, we present hardware and software tools that allow other researchers to reproduce our system and results. 
For videos and additional details, please see \href{https://sites.google.com/nvidia.com/industreal}{our project website}. 
\end{abstract}

\IEEEpeerreviewmaketitle

\section{Introduction}
\label{sec:introduction}
Robotic assembly is a longstanding challenge \cite{whitney_mechanical_2004, kimble2022performance}. 
Assembly requires contact-rich interactions and high precision and accuracy; high-mix, low-volume settings also require adaptivity to diverse parts, poses, and environments. 
Today, robotic assembly is ubiquitous in the automotive, aerospace, and electronics industries. 
However, assembly robots are typically expensive, achieving precision primarily through hardware rather than intelligence.
Moreover, systems often require meticulous engineering of adapters, fixtures, lighting, and robot trajectories. 
Such efforts demand substantial time and effort from robotics integrators and can result in solutions that are highly sensitive to perturbations of the robotic workcell. 

Simulation is an indispensable means to solve engineering challenges. For example, simulators are used for finite element analysis, computational fluid dynamics, and integrated circuit design. 
Nevertheless, simulation has had a comparably limited impact on robotic assembly. 
Accurate simulation of geometrically-complex parts can require generating 1$k$-10$k$ contacts per rigid body pair, followed by solving a nonlinear complementarity problem (NCP) at each contact. 
Furthermore, robotic assembly tasks require long-horizon sequential decision-making; powerful data-driven methods (e.g., on-policy RL) have high sample complexity, requiring faster-than-realtime simulation.
Achieving such accuracy and performance requirements has only recently become possible \cite{macklin_local_2020, narang2022factory, lan2022affine}. 

\begin{figure}
    \includegraphics[width=\columnwidth]{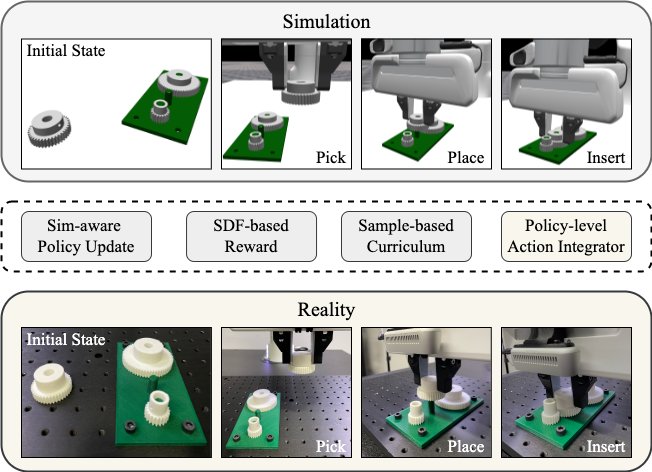}
    \caption{\footnotesize \textbf{Overview.} Top: Simulation-based policy learning for one of our tasks, gear assembly. Middle: Proposed algorithms to facilitate sim-based learning and real-world deployment. Bottom: Successful transfer to the real world.}
    \label{fig:teaser}
\end{figure}

Given modeling limitations and finite compute, simulation will always differ from reality; this reality gap has been notoriously large for robotics.
Sim-to-real transfer methods address this gap through techniques such as system identification and domain randomization.
These methods have shown remarkable results in locomotion \cite{lee_learning_2020, rudin_learning_2021, margolis2022rapid} and manipulation \cite{andrychowicz_learning_2020, handa_dextreme_2022, chen2022visual}. However, sim-to-real efforts for assembly have been scarce.

We present \textbf{IndustReal}, a set of algorithms, systems, and tools for solving contact-rich assembly tasks in simulation and transferring behaviors to reality (\autoref{fig:teaser}). Specifically, our primary contributions are the following:

\begin{itemize}
    \item \textbf{Algorithms}: For simulation, we propose three methods to allow RL agents to solve contact-rich tasks in a simulator: a \textbf{simulation-aware policy update (SAPU)} to reward the agent when simulation predictions are more reliable, a \textbf{signed distance field (SDF)-based dense reward} to provide an alignment metric between geometrically-complex objects, and a \textbf{sampling-based curriculum (SBC)} to prevent overfitting to earlier stages of a curriculum. For sim-to-real transfer, we also propose a \textbf{policy-level action integrator (PLAI)}, which reduces steady-state error in the presence of unmodeled dynamics (e.g., friction).
    \item \textbf{Benchmarks}: We solve several challenging tasks proposed in Factory \cite{narang2022factory} (pick, place, and insertion tasks for pegs-and-holes and gear assemblies in simulation) with success rates of 82-99\%. We provide careful evaluations over 265k simulated trials to show the utility of SAPU, SDF-based rewards, and SBC for solving these tasks.
    \item \textbf{Systems}: We design and demonstrate a real-world system that can perform sim-to-real transfer of our simulation-trained policies, with success rates of 83-99\% over 600 trials. We provide careful evaluations to show the utility of PLAI. To our knowledge, this is the first system for sim-to-real of all phases of the assembly problem: from detection, to grasping, to part alignment, to insertion. Our system uses commonly-used robotics hardware and requires no real-world policy adaptation phase.
\end{itemize}

Our secondary contributions are the following:
\begin{itemize}
    \item \textbf{Hardware}: We present \textbf{IndustRealKit}, 
    which contains CAD models for all parts designed for our setup, as well as a list of all purchased parts. The CAD models can all be printed on a desktop 3D printer. IndustRealKit allows the research community to easily replicate our experimental hardware and benchmark their performance.
    \item \textbf{Software}: We present \textbf{IndustRealLib}, a lightweight Python library that allows users to easily deploy policies trained in NVIDIA Isaac Gym \cite{makoviychuk_isaac_2021} onto a real-world Franka Emika Panda robot \cite{haddadin2022franka}. The library also contains code to assist with policy training. %
    IndustRealLib allows the research community to reproduce our robot behaviors.
\end{itemize}
We aim for \textbf{IndustReal} to provide algorithms, benchmark results, and a reproducible system that serve as a path forward for sim-to-real transfer on contact-rich assembly tasks.

\section{Related Work}
\label{sec:related-work}

We divide prior work on robotic assembly into three categories: 1) classical approaches leveraging analytical methods \cite{whitney_mechanical_2004, mason2001mechanics}, 2) learning-based approaches leveraging real-world data or experience \cite{xu2019compare}, and 3) RL-based sim-to-real approaches leveraging robotics simulators. We defer a review of (1) and (2) to Appendix~\ref{sec:appendix_related_work} and focus on (3).

\subsection{Sim-to-Real Transfer for Assembly}
\label{subsec:related-transfer}

Over the past few years, there have been a number of impressive efforts in sim-to-real for assembly. These efforts have primarily used MuJoCo \cite{schoettler2020meta, davchev_residual_2021, hebecker_towards_2021, vuong_learning_2021, zhang_learning_2021, kozlovsky2022reinforcement} or PyBullet \cite{luo_dynamic_2019, son_sim-to-real_2020, shao_learning_2020}; have used PPO \cite{son_sim-to-real_2020, spector_deep_2020, hebecker_towards_2021, kozlovsky2022reinforcement} or DDPG \cite{luo_dynamic_2019, apolinarska_robotic_2021}; and have aimed to solve peg-in-hole \cite{luo_dynamic_2019, beltran-hernandez_variable_2020, shao_learning_2020, schoettler2020meta, spector_deep_2020, davchev_residual_2021, vuong_learning_2021, zhang_learning_2021, kozlovsky2022reinforcement} or NIST-style tasks \cite{beltran-hernandez_variable_2020, schoettler2020meta, davchev_residual_2021, zhang_learning_2021}. However, several of these studies use large clearances (e.g., $\geq 1~mm$) and/or large parts in simulation and/or the real world.
Furthermore, almost all use force/torque (F/T) sensors to collect observations and/or set thresholds. Most require human demonstrations \cite{luo_dynamic_2019, apolinarska_robotic_2021, davchev_residual_2021}, a baseline motion plan \cite{son_sim-to-real_2020, spector_deep_2020, kozlovsky2022reinforcement}, and/or fine-tuning in the real-world \cite{beltran-hernandez_variable_2020, schoettler2020meta}. Finally, all but one \cite{schoettler2020meta} focus only on insertion and assume the object is pre-grasped; 
however, \cite{schoettler2020meta} also uses specialized grippers, low-dimensional action spaces, highly-constrained target locations, and a real-world policy adaptation phase.

In contrast, we make several design choices that increase the realism of the problem and encourage reproducibility.
First, for software, we use Factory \cite{narang2022factory} within Isaac Gym, which can solve contact dynamics between highly-complex geometries without simplification. Second, for hardware, we use a collaborative robot (Franka Panda) and RGB-D camera (Intel RealSense D435) that are widespread in research, but far less precise than those in industrial assembly. We use no task-specific grippers and avoid F/T sensors due to their cost, noise, %
and fragility. We also use realistic part clearances ($\leq $0.5-0.6 mm, aligned with the upper bound of ISO 286). Next, for problem scope, we address sim-to-real for all parts of the assembly sequence (i.e., detection, grasping, alignment, and insertion). We face robustness challenges due to calibration and localization error; moreover, we apply large randomizations of part poses and targets. Finally, in methodology, we achieve sim-to-real transfer without baseline plans or demos, dynamics randomization, or real-world policy adaptation phases.

\section{Problem Description}
\label{sec:method-problem}

\subsection{Problem Setup}
Our problem setup is as follows: a Franka robot is mounted to a work surface. 
A RealSense D435 camera is mounted to the robot wrist. Industrial-style parts inspired by the NIST Task Board 1 \cite{kimble_benchmarking_2020, kimble2022performance} are placed upright on the work surface. 
The parts with extruded features (which we henceforth refer to as \textit{plugs}) have a randomized 3-DOF pose ($x$, $y$, $\theta$) on top of an optical breadboard and are free to move; the parts with mating features (which we call \textit{sockets}) also have a randomized pose, but are bolted to the breadboard to emulate industrial fixturing.\footnote{This broad usage of \textit{plug} and \textit{socket} will persist throughout the paper.} The fundamental task is to perceive, grasp, transport, and insert all the plugs into their corresponding sockets.

\begin{figure}
    \includegraphics[width=\columnwidth]{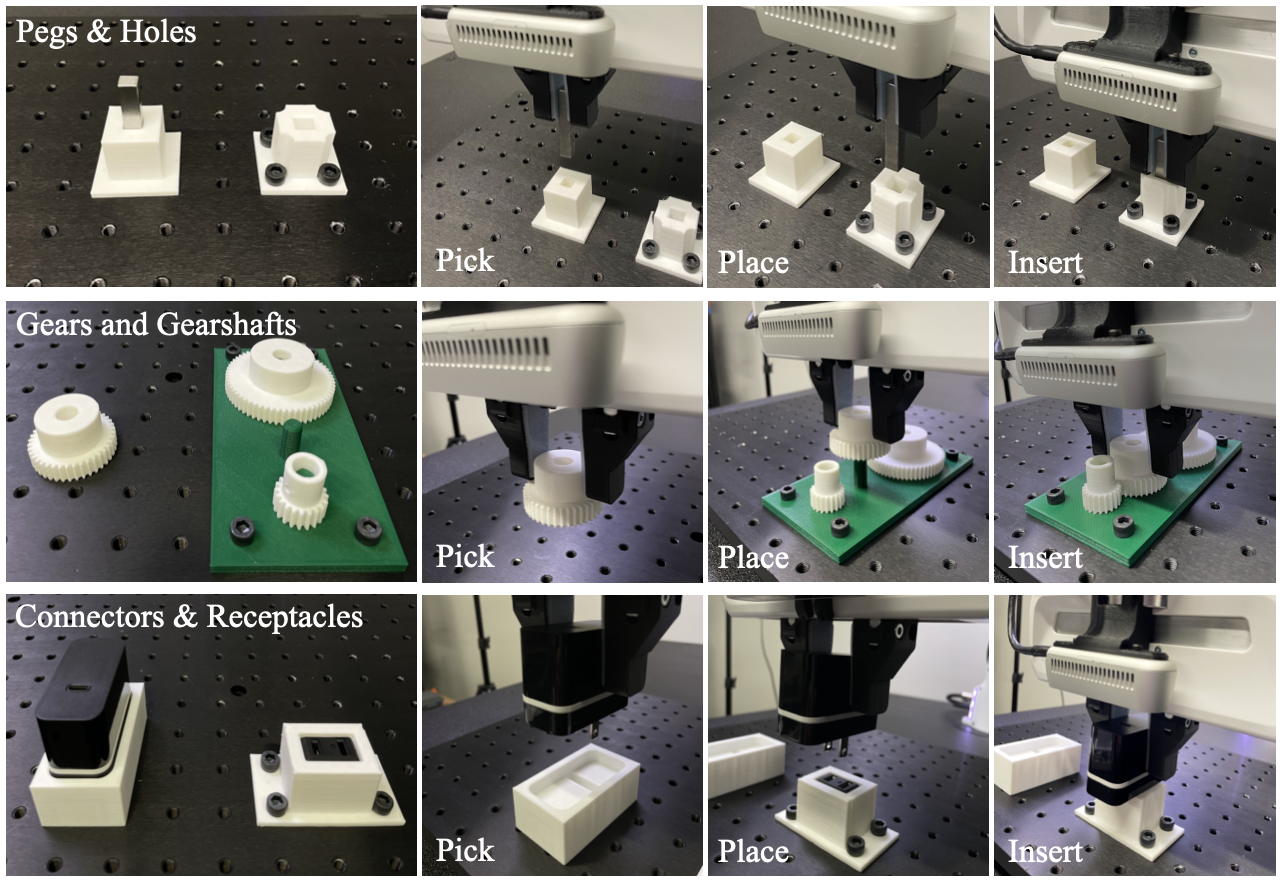}\caption{\footnotesize{\textbf{Problem setup and decomposition.} Column 1: Three types of assemblies. Columns 2-4: Goal states of Pick, Place, and Insert phases.}}
    \label{fig:formulation}
\end{figure}

Specifically, we aim to perform this task for three types of assemblies from \cite{narang2022factory} (\autoref{fig:formulation} column 1):
\begin{itemize}
    \item \textbf{Pegs and Holes:} 2 different classes of pegs (round and rectangular), each with 3 different sizes (max dimension: 8 mm, 12 mm, and 16 mm) must be inserted into corresponding holes (clearances: 0.5-0.6 mm).
    \item \textbf{Gears and Gearshafts:} 3 different gears (diameters: 20 mm, 40 mm, 60 mm) must be inserted onto corresponding gearshafts (diametral clearances: 0.5 mm).
    \item \textbf{Connectors and Receptacles:} 2 different connectors (2-prong NEMA 1-15P and 3-prong NEMA 5-15P) must be inserted into corresponding receptacles.
\end{itemize}

We first aim to solve the assembly task in simulation using RL with CAD models of the robot and the objects. We then aim to transfer the policies to the real world. 

\subsection{Problem Decomposition}

For each category of parts, to facilitate the assembly task, we decompose it into three phases (\autoref{fig:formulation} columns 2-4):
\begin{itemize}
    \item \textbf{Pick}: The robot grasps a randomly-positioned plug (i.e., peg, gear, or connector) within its workspace.
    \item \textbf{Place}: The robot transports the grasped plug close to its corresponding socket (i.e., hole, gearshaft, or receptacle).
    \item \textbf{Insert}: The robot brings the grasped plug into contact with its socket and inserts the plug, aligning parts where necessary (e.g., when inserting intermediate gears).
\end{itemize}

\section{Policy Learning in Simulation}
\label{sec:method-policy-learning-general}

In this section, we first describe our general strategies for policy learning in Sections~\ref{subsec:training-environments}-\ref{subsec:randomization-and-noise}. Although these strategies were sufficient to train successful \textbf{Pick} and \textbf{Place} policies, they were inadequate for training \textbf{Insert} policies ($\approx 12\%$ success rates), motivating our algorithmic development. We describe our 3 simulation-based algorithms \textit{and evaluate them} on our \textbf{Insert} policies
in Sections~\ref{subsec:simulation-aware-policy-update}-\ref{subsec:curriculum-learning}\footnote{Note that when we evaluate each algorithm, the other two algorithms are used; thus, the evaluations function as ablation studies.}. 

\subsection{Training Environments}
\label{subsec:training-environments}

We developed our code within the Factory simulation framework \cite{narang2022factory}. For the \textbf{Peg and Hole} assemblies, we first trained a \textbf{Reach} policy, where the robot learned to move its end-effector to a randomized pose within a large workspace. We then fine-tuned \textbf{Reach} to solve the \textbf{Pick} task by including all peg assets in the scene and redefining success as lifting the pegs. Similarly, we fine-tuned \textbf{Reach} to solve the \textbf{Place} task by initializing the pegs within the robot grippers, including all hole assets in the scene, and redefining success as bringing the pegs to their corresponding holes. Empirically, for the \textbf{Pick} and \textbf{Place} tasks, training in free space and fine-tuning on contact was more efficient than training from scratch. However, we trained an \textbf{Insert} policy from scratch.

For the \textbf{Gears and Gearshafts} assemblies, we did not train policies to solve the \textbf{Pick} or \textbf{Place} tasks; as we later show, we solved those tasks in the real world by executing the corresponding \textbf{Peg and Hole} policies, demonstrating generalization. However, we again trained an \textbf{Insert} policy from scratch.

Finally, for the \textbf{Connectors and Receptacles} assemblies, we did not train policies for any phase. Again, we later show that we solved those tasks in the real world via policy transfer.

In summary, for the \textbf{Peg and Hole} assemblies, we trained \textbf{Pick}, \textbf{Place}, and \textbf{Insert} policies, and for the \textbf{Gears and Gearshafts} assemblies, we trained another \textbf{Insert} policy.

\subsection{Formulation}

We formulated the problem as a Markov decision process (MDP) with state space $\mathcal{S}$, observation space $\mathcal{O}$, action space $\mathcal{A}$, state transition dynamics $\mathcal{T}: \mathcal{S} \times \mathcal{A} \rightarrow \mathcal{S}$, initial state distribution $\rho_0$, reward function $r: \mathcal{S}  \rightarrow \mathbb{R}$, horizon length $T$, and discount factor $\gamma \in (0, 1]$. The objective was to learn a policy $\pi : \mathcal{O} \rightarrow \mathbb{P}(\mathcal{A})$ that maximized the expected sum of discounted rewards $\mathbb{E}_{\pi}[\Sigma^{T-1}_{t=0}\gamma^t r(s_t)]$.

We used proximal policy optimization (PPO) \cite{schulman2017proximal} 
to learn a stochastic policy $a \sim \pi_{\theta} (o)$ (actor), mapping from observations $o \in \mathcal{O}$ to actions $a \in \mathcal{A}$ and parameterized by a network with weights $\theta$; as well as an approximation of the on-policy value function $v = V_{\phi} (s)$ (critic), mapping from states $s \in \mathcal{S}$ to value $v \in \mathcal{V}$ and parameterized by  weights $\phi$. 
We used the PPO implementation from \textit{rl-games} \cite{rl-games2022}; %
hyperparameters and architectures are in \autoref{tab:ppo-param}. Finally, we aimed to train the policy in simulation and deploy in the real world with no policy adaptation phase on the specific real environment.

\subsection{Observations, Actions, and Rewards}
Our observation spaces in simulation and the real world were task-dependent. The observations provided to the actor consisted exclusively of joint angles, gripper/object poses, and/or target poses, as the Franka's joint velocities and joint torques exhibited appreciable noise in the real world. However, we employed asymmetric actor-critic \cite{pinto2017asymmetric}, where velocity information was still used to train the critic. Our exact observations for all policies are listed in \autoref{tab:method-observation}.

Our action spaces for both simulation and the real world were task-independent. The actions consisted of incremental pose targets to a task-space impedance (TSI) controller (specifically, $a = [\Delta x; \Delta q]$, where $\Delta x$ is a position error and $\Delta q$ is a quaternion error). 
We learned incremental targets rather than absolute targets because the latter encodes task-specific biases and must be selected from a large spatial range. We used TSI rather than operational-space control (OSC) because Franka provides a high-performance implementation of TSI, and OSC relies on an accurate dynamics model.

Our rewards in simulation were task-dependent. However, all rewards could be expressed in the following general form:
\begin{multline}
    G = w_{h_0}..w_{h_m} \bigg( \sum_{t=0}^{H-1} [w_{d_0} R_{d_0} (t) + ... + w_{d_n} R_{d_n} (t)] \\ + w_{s_0} R_{s_0} + ... + w_{s_p} R_{s_p} \bigg)
\end{multline}
where $G$ is the return over the horizon, ${R_d}_0...{R_d}_n$ are distinct dense rewards, 
$H$ is the horizon length, 
$R_{s_0}...R_{s_p}$ are terminal success bonuses,
$w_{d_0}...w_{d_n}$ and $w_{s_0}...w_{s_p}$ are scaling factors that map distinct rewards into a consistent unit system 
and weight the importance of each term, and $w_{h_0}...w_{h_m}$ are scaling factors on the return over the entire horizon. Not all terms are used in each phase, and most of our reward formulations are simple. Detailed formulations and success criteria are provided in \autoref{tab:method-reward} and \autoref{tab:method-task-success}, respectively.

\subsection{Randomization and Noise}
\label{subsec:randomization-and-noise}

At the start of each episode, we randomized the 6-DOF end-effector and object poses over a large spatial range. In addition, for the \textbf{Insert} policies, we introduced observation noise. As well established, these perturbations are critical for ensuring robustness to initial conditions and sensor noise in the real world. Randomization and noise ranges are provided in \autoref{tab:sim-randomization}). However, we avoided dynamics randomization \cite{peng2018sim, handa_dextreme_2022}, as we had strong priors on our system dynamics.

\subsection{Simulation-Aware Policy Update (SAPU)}
\label{subsec:simulation-aware-policy-update}

\textbf{Method:} In contact-rich simulators, spurious interpenetrations between assets are unavoidable, especially when executing in real-time (Figure~\ref{fig:sim_ip_checking}). %
Unfortunately, in simulation for RL, an agent can exploit inaccurate collision dynamics to maximize reward, learning policies that are unlikely to transfer to the real world \cite{muratore2019assessing}. 
Thus, we propose our first algorithm, a \textbf{simulation-aware policy update (SAPU)}, where the agent is encouraged to learn policies that avoid interpenetrations.

Specifically, we implemented a GPU-based interpenetration-checking module using \textit{warp} \cite{warp2022}.
For a given environment, the module takes as input the plug and socket mesh and associated 6-DOF poses. %
The module samples $N=1000$ points on/inside the mesh of the plug, transforms the points to the socket frame, computes distances to the socket mesh, and returns the max interpenetration depth (\autoref{alg:interpenetration-checking}).
This procedure is performed each episode, and the depth is used to weight the cumulative reward during the policy update. 

\textbf{Evaluation:} We evaluate \textbf{SAPU} on the \textbf{Peg and Hole} assembly \textbf{Insert} policy, with the following test cases:
\begin{itemize}
    \item \textbf{Baseline:} Do not utilize interpenetration information.
    \item \textbf{Filter Only:} For a given episode, if max interpenetration depth $d_{ip}^{max}$ is greater than $\epsilon_{ip} = 1~mm$, do not use return in policy update. If $d_{ip}^{max} < \epsilon_{ip}$, use return as normal.
    \item \textbf{Weight Only:} For a given episode, weight return by $1-\tanh(d_{ip}^{max}/\epsilon_{d})$, which is bounded by $(0, 1]$.
    \item \textbf{Filter and Weight:} For a given episode, if $d_{ip}^{max} > \epsilon_{ip}$, do not use return in policy update. If $d_{ip}^{max} < \epsilon_{ip}$, weight return by $1-\tanh(d_{ip}^{max}/\epsilon_{d})$).
\end{itemize}

\begin{figure}
    \centering
    \includegraphics[width=0.48\textwidth]{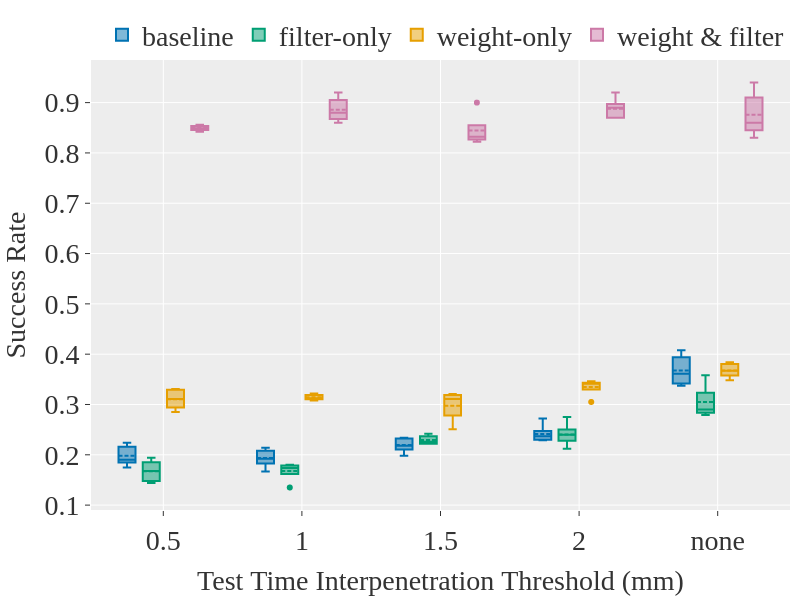}
    \caption{\footnotesize{\textbf{Evaluation of Simulation-Aware Policy Update}. Success rates are computed for episodes where the maximum interpenetration distance was less than the specified value at test time. Boxes indicate median and IQR.}}
    \label{fig:sim_ablation_ip}
\end{figure}

After training policies with each strategy, we tested each policy in simulation over 5 seeds, with 1000 trials per seed; quantified $d_{ip}^{max}$ for each episode; and evaluated success rate over all episodes (\autoref{fig:sim_ablation_ip}). Success was defined as inserting the peg into the hole (\autoref{tab:method-task-success}); however, to also ascertain whether success was achieved in the desired way (i.e., by avoiding interpenetration), we computed success rate for episodes where $d_{ip}^{max} < [0.5, 1, 1.5, 2, \infty]$ mm. %

For both the most realistic scenario (when only successes where $d_{ip}^{max} < 0.5$ mm were counted) and the least realistic scenario (when all successes were counted), the \textbf{Baseline}, \textbf{Filter Only}, and \textbf{Weight Only} strategies were unable to achieve success rates above 40\%. However, the \textbf{Filter and Weight} strategy performed well over all scenarios, with success rates of $84.9$- $87.6\%$. Thus, \textbf{Filter and Weight} was not only most effective at learning a policy that avoided interpenetration, but was also most effective at policy learning \textit{in general}.

\begin{algorithm}
    \SetAlFnt{\small}
    \KwIn{plug mesh $m_p$, socket mesh $m_s$, plug pose $p_p$, socket pose $p_s$, number of query points $N$.}
    sample N points in $m_p \rightarrow \mathbf{v}$, $\mathbf{v}=\{v_0, ..., v_{N-1}\}$\;
    transform $\mathbf{v}$ to current $m_p$ pose $p_p$ in $m_s$ frame \;
    \For{every vertex $v \in \mathbf{v}$}{
        compute closest point on $m_s$ to $v$\;
        \If{$v$ inside $m_s$}{
            calculate interpenetration distance\;
        }
    } 
    $d_{ip}^{max}=$ max interpenetration from all $v\in \mathbf{v}$ to $m_s$\;
    \Return $d_{ip}^{max}$\;
    \caption{Interpenetration Checking Per Env.}
    \label{alg:interpenetration-checking}
\end{algorithm}

\subsection{SDF-Based Dense Reward}
\label{subsec:sdf-based-dense-reward}

\textbf{Method}: Keypoint-based rewards are widely used, as they avoid weighting between distinct position and orientation rewards \cite{allshire2021transferring}. However, collinear keypoints (e.g., \cite{narang2022factory}) underspecify the assembly of non-axisymmetric parts (e.g., rectangular pegs), and non-collinear keypoints overspecify the assembly of symmetric parts, as identical geometries do not alias. Thus, we propose a \textbf{signed distance field (SDF)-based dense reward}, where an SDF is defined as a map $\phi(\boldsymbol{x}): \mathbb{R}^3 \rightarrow \mathbb{R}$ from an arbitrary point $\boldsymbol{x}$ to its signed distance $\phi(\boldsymbol{x})$ to a surface.

Specifically, for each plug mesh in its nominal pose, we use sampling to preselect $N=1000$ points on the surface. In addition, for the same mesh in its \textit{target pose}, we use \textit{pysdf} to precompute and store the SDF values at each cell of a dense voxel grid containing the mesh. During training, the preselected points are transformed to the frame of the plug mesh in its \textit{current pose}, and the SDF values are queried at these points. 
This procedure is performed at each timestep in each environment and is used to generate a reward signal.

\textbf{Evaluation:} We evaluate \textbf{SDF-Based Dense Reward} on the \textbf{Pegs and Holes} assembly \textbf{Insert} policy by comparing the following reward formulations (\autoref{tab:sim-ablation-rwd}):
\begin{itemize}
    \item \textbf{Collinear Keypoints:} Each object has 4 keypoints along its Z-axis; Euclidean distances are summed and averaged.
    \item \textbf{6-DOF Keypoints:} Each object has 13 keypoints over 3 axes; Euclidean distances are summed and averaged.
    \item \textbf{Chamfer Distance:} Each object has a point cloud defined by its mesh vertices; chamfer distance \cite{fan_learning_2019} is computed.
    \item \textbf{SDF Query Distance:} The root-mean-square SDF distance is computed as described earlier.
\end{itemize}

\begin{table*}[t]
\centering
\rowcolors{2}{white}{Gainsboro} 
\bgroup
\def\arraystretch{1.2}%
\begin{tabular}{ lcccccc } 
\toprule
Reward & Formulation & Num. Trials & Success  (\%) & Engage (\%) & Pos. Error (mm) & Rot. Error (rad.) \\ \midrule
Collinear keypoints & $-||\mathbf{k}_{\mathrm{curr}}-\mathbf{k}_{\mathrm{goal}}||^2$  & 1000 & 15.40 $\pm$ 5.22 & 64.40 $\pm$ 3.05 & 16.28 $\pm$ 1.08 & 0.150 $\pm$ 0.020 \\
6-DOF keypoints & $-||\mathbf{k}^{6D}_{\mathrm{curr}}-\mathbf{k}^{6D}_{\mathrm{goal}}||^2$  & 1000 & 54.2 $\pm$ 7.56 & 83.80 $\pm$ 4.44 & 11.74 $\pm$ 1.92 & 0.132 $\pm$ 0.013\\
Chamfer distance & $-\mathrm{Chamfer\_dist(S_{\mathrm{plug}}, S_{\mathrm{socket}})} $  & 1000 & 1.80 $\pm$ 1.92 & 47.80 $\pm$ 6.14 & 31.02 $\pm$ 0.95 & 0.994 $\pm$ 0.030 \\
SDF query distance & $-\log(\Sigma_{i}^{N}\phi(x_i)/N)$ & 1000 & 88.60 $\pm$ 2.41 & 96.60 $\pm$ 2.30 & 3.80 $\pm$ 0.80 & 0.086 $\pm$ 0.026 \\
\bottomrule
\end{tabular}
\egroup
\caption{\footnotesize{\textbf{Evaluation of SDF-Based Dense Reward.}} Symbol $\mathbf{k}$ denotes object keypoint positions, $S$ is a set of points comprising a point cloud (here, we use plug/socket mesh vertices), and $x_i$ denotes points sampled from the plug mesh (again, we use vertices). \textit{Engage} denotes a partial insertion.}
\label{tab:sim-ablation-rwd}
\end{table*}

After training policies with each strategy, we tested each policy in simulation over 5 seeds, with 1000 trials per seed; quantified terminal position and rotation error for each episode; and quantified success rate and engagement rate (\autoref{tab:sim-ablation-rwd}). Success was defined as inserting the peg into the hole; \textit{engagement} was defined as a partial insertion.

The \textbf{Collinear Keypoints}, \textbf{6-DOF Keypoints}, and \textbf{Chamfer Distance} rewards resulted in appreciable position and rotation errors of 11.7-31.0 mm and 0.13-0.99 rad, respectively, with chamfer distance performing the worst; as follows, success rates varied between 1.8-54.2\%. However, the \textbf{SDF Query Distance} reward resulted in errors of just 3.80 mm and 0.086 rad, with a success rate of 88.6\% and near-perfect engagement rate of 96.6\%. Thus, \textbf{SDF Query Distance} was by far the most effective reward formulation for policy learning. As Factory \cite{narang2022factory} already precomputes SDFs for all objects for contact generation, we envision a single representation-generation step for both physics and reward.

\subsection{Sampling-Based Curriculum (SBC)}
\label{subsec:curriculum-learning}

\textbf{Method:} 
Curriculum learning \cite{bengio2009curriculum} is an established approach for solving long-horizon problems; as the agent learns, the difficulty of the task is gradually increased. Nevertheless, for both the \textbf{Peg and Hole} and \textbf{Gears and Gearshafts} assemblies, for the \textbf{Insert} phase, naive implementations of curriculum learning (i.e., increasing initial distance from goal) were ineffective; when the initial peg/gear state was above the hole/gearshafts, the agent failed to progress, likely overfitting to a partially-inserted plug. Thus, we developed \textbf{Sampling-Based Curriculum (SBC)}, whereby the agent is exposed to the entire range of initial state distributions from the start of the curriculum, but the lower bound is increased at each stage.

Specifically, let $z^{low}$ denote the lower bound of the initial height of a plug above its socket at a given curriculum stage, and let $z^{high}$ denote a constant upper bound; the initial height of the plug is uniformly sampled from $\mathrm{Uniform}[{z^{low}}, z^{high}]$.
In addition, let $\Delta z^i$ and $\Delta z^d$ denote an increase or decrease in $z^{low}$, and let
$p_n$ denote the mean success rate over all environments during episode $n$.
When episode $n$ terminates, we update ${z^{low}}$ as follows:
\begin{align*}
{z^{low}} \leftarrow
\left\{\begin{matrix}
z^{low} + \Delta z^i, & p_n>80\%\\ 
z^{low} - \Delta z^d, & p_n<10\%\\
z^{low}, & \text{otherwise.}
\end{matrix}\right.
\end{align*}

In general, we enforce $\Delta z^d<\Delta z^i$. We define an increase in $z^{low}$ as an advance to the next stage of the curriculum, and a decrease in
$z^{low}$ as a reversion to the previous stage.

\textbf{Evaluation:} We evaluate \textbf{SBC} on the \textbf{Pegs and Holes} assembly \textbf{Insert} phase, with the following test cases:

\begin{itemize}
    \item \textbf{Baseline:} No curriculum learning is used; $z^{init} = z^{high}$.
    \item \textbf{Standard:} Peg height is initialized at ${z^{low}}$; at each stage, ${z^{low}}$ increases, until a max value of ${z^{high}}$.
    \item \textbf{Sampling-Based:} Initial peg height is sampled as described earlier.
\end{itemize}

For the \textbf{Standard} and \textbf{Sampling-Based} strategies, $z_{low}$ was initially 10 mm below the top of the hole, and $z_{high}$ remained constant at 10 mm above. The criterion for advancing to the next stage was an 80\% success rate. \textit{Engagement} was defined as partially inserting the peg into the hole; success was defined as full insertion. We set $\Delta z^i = 5 mm$ and $\Delta z^d = 3 mm$. Whenever the peg was initialized above the hole, we also perturbed its position along the X- and Y-axes.

After training policies with each strategy, we tested each policy in simulation over 5 seeds, with 1000 trials per seed (\autoref{tab:sim_ablation_curriculum}). All strategies achieved moderate rotation errors of 0.086-0.096 rad, and the \textbf{Baseline} and \textbf{Sampling-Based} strategies both achieved high engagement rates of 89.2-96.6\%. However, \textbf{Sampling-Based} substantially outperformed the others in success rate  and position error; it achieved 88.6\% and 3.80 mm, respectively, whereas the others performed no better than 66.8\% and 10.7 mm.
These results substantiate existing evidence that curriculums can facilitate RL when carefully implemented, and importantly, suggest a specific implementation in the case of discontinuous contact.

\begin{table}[]
\centering
\rowcolors{2}{white}{Gainsboro}
\scalebox{0.85}{\begin{tabular}{ lcccc } 
\toprule
\textbf{Curriculum} & \textbf{Success (\%)} & \textbf{Engage (\%)} & \textbf{Pos. Err.} (mm) & \textbf{Rot. Err.} (rad)\\ \midrule
Baseline & 66.80 $\pm$ 5.76 & 89.2 $\pm$ 4.97 & 10.70 $\pm$ 2.70 & 0.086 $\pm$ 0.095\\
Standard & 32.40 $\pm$ 1.82 & 46.0 $\pm$ 3.61 & 18.16 $\pm$ 0.36 & 0.096 $\pm$ 0.0091 \\
Sampling-Based & 88.60 $\pm$ 2.41 & 96.6 $\pm$ 2.30 & 3.80 $\pm$ 0.80 & 0.086 $\pm$ 0.026 \\
\bottomrule
\end{tabular}}
    \caption{\footnotesize \textbf{Evaluation of Sampling-Based Curriculum.} \textit{Baseline} denotes that no curriculum was used.}
    \label{tab:sim_ablation_curriculum}
\end{table}

\subsection{Joint Evaluation}
\begin{figure*}
\centering
    \includegraphics[width=\textwidth]{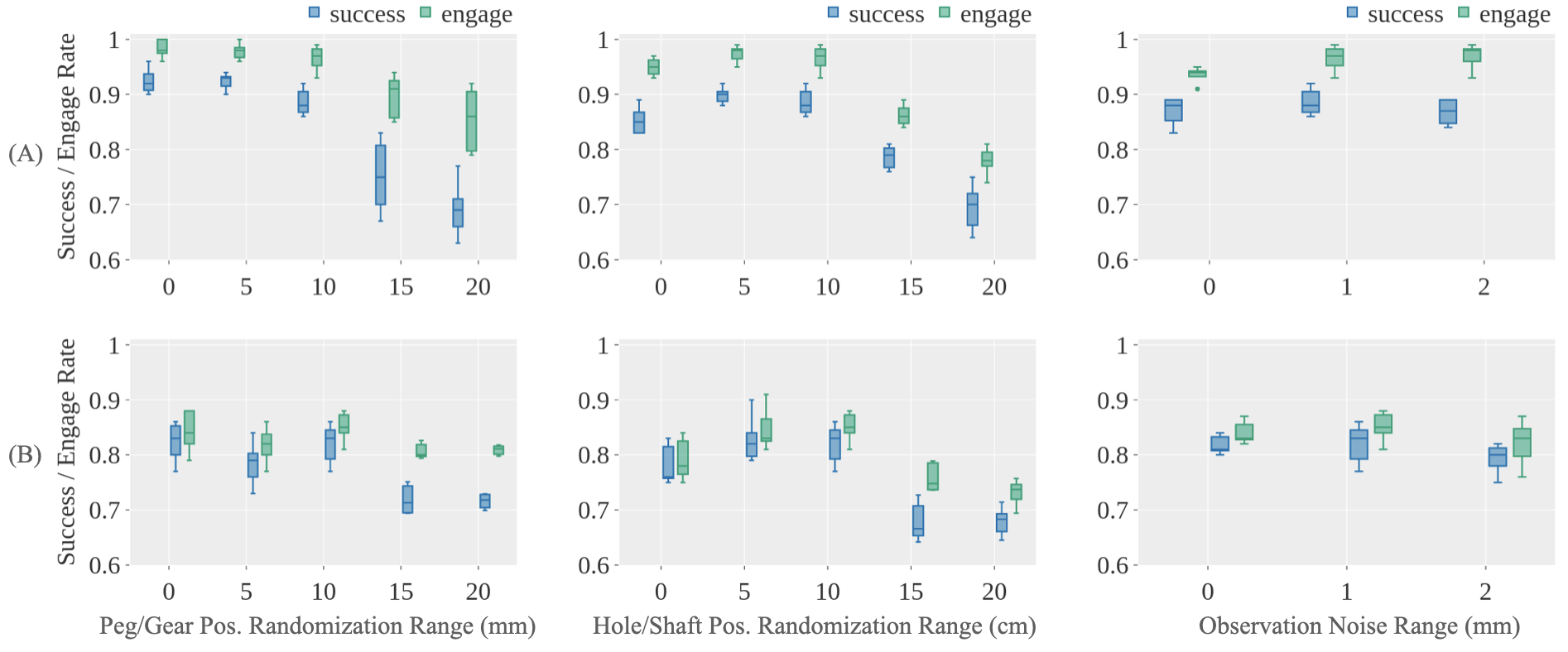}
\caption{\footnotesize \textbf{Joint evaluation of Simulation-Based Policy Update, SDF-Based Dense Reward, and Sampling-Based Curriculum.} (A) \textbf{Pegs and Holes} assembly \textbf{Insert} policy. (B) \textbf{Gears and Gearshafts} assembly \textbf{Insert} policy. \textit{Engage} denotes partial insertion. Policies were trained with moderate randomization (plug/socket randomization of $\pm$ 10 mm and 10 cm, respectively, and observation noise of $\pm$ 1 mm); thus, plots evaluate in-distribution and OOD performance.}
\label{fig:sim-eval-insertion}
\end{figure*}

As described in Sections~\ref{subsec:simulation-aware-policy-update}-\ref{subsec:curriculum-learning}, we proposed three algorithms for improving learning of contact-rich \textbf{Insert} policies: \textbf{Simulation-Aware Policy Update} to adapt to simulator inaccuracy, \textbf{SDF-Based Dense Reward} to quantify alignment for asymmetric or symmetric objects, and \textbf{Sampling-Based Curriculum} to prevent overfitting to initial partial insertions. As a final evaluation, we comprehensively evaluated all three techniques in tandem (\autoref{fig:sim-eval-insertion} and \autoref{tab:sim_eval_peg_gear}). When training and testing with moderate state randomization (plug/hole randomization of $\pm$10 mm/$\pm$10 cm, respectively) and observation noise ($\pm 1$ mm), the \textbf{Pegs and Holes} assembly \textbf{Insert} policy achieved success and engagement rates of 88.6\% and 96.6\%, respectively, whereas the \textbf{Gears and Gearshafts} assembly \textbf{Insert} policy achieved 82.0\% and 85.2\%. Across all evaluations, worst-case performance was 67.88\% (when testing with \textit{twice} the gearshaft position randomization of training), and best-case was 92.4\% (testing with no randomization or noise).

\section{Policy Deployment in Real World}
\label{subsec:method-perception}

In this section, we first describe our general strategies for policy deployment in the real world in Sections~\ref{sec:communications-framework}-\ref{sec:dynamics-strategy}. We then describe \textit{and evaluate} our deployment-time algorithm, the \textbf{Policy-Level Action Integrator (PLAI)}, in Section~\ref{sec:policy-level-action-integrator}. Finally, we provide comprehensive evaluations and demonstrations of our full real-world system in Section~\ref{sec: eval-insertion-real}.

\subsection{Communications Framework}
\label{sec:communications-framework}

We illustrate our communications stack in Figure~\ref{fig:communications}. In summary, we developed the \textbf{IndustRealLib} library, which accepts trained policy checkpoints from Isaac Gym as input, and outputs targets for a Franka robot controlled via a task-space impedance (TSI) controller. The targets are sent to the \textit{frankapy} Python library, which streams the commands via ROS to the \textit{franka-interface} C++ library \cite{zhang2020modular}. \textit{franka-interface} then sends the commands to the \textit{libfranka} library provided by Franka, which maps the TSI commands to low-level joint torques that are streamed to the robot. Proprioceptive signals are then communicated to \textbf{IndustRealLib} in reverse order.

Latencies of our real-world and simulated systems are summarized in \autoref{tab:latency}. In the real world, physics frequency is approx. infinite, low-level control frequencies (between \textit{libfranka} and robot) are 1 kHz, and policy control frequencies (between \textit{libfranka} and \textbf{IndustRealLib}) are 60-100 Hz (limited by inference and ROS). We thus set our physics frequency during training to the highest practical rate (120 Hz) given our compute, and restricted our control rate during training to 60 Hz to prevent aliasing of policy signals during deployment.

\subsection{Perception Pipeline}
\label{sec:perception-pipeline}

The primary goal of our perception pipeline is to estimate the 2D poses ($x$, $y$, $\theta$) of the parts in the robot frame. Our pipeline consists of 3 separate steps: a one-time camera calibration, a per-experiment workspace mapping, and a per-experiment object detection from a single RGB image. Bounding box centroids and a trivially-specified height are used to construct targets. All details are provided in Appendix~\ref{appendix:camera-calibration}-\ref{appendix:constructing-targets}.

\subsection{Dynamics Strategy}
\label{sec:dynamics-strategy}

Our policies initially exhibited substantial steady-state error during deployment. We note that a partial resolution was carefully eliminating all arbitrary energy dissipation (e.g., heuristically-applied friction and damping) in the simulator and asset descriptions; dissipative terms are often introduced to facilitate simulation stability, but their values are typically chosen without physical consideration. Simulation and real-world results with/without heuristic damping are in \autoref{fig:eval_place}.

\subsection{Policy-Level Action Integrator}
\label{sec:policy-level-action-integrator}

\textbf{Method:} Robotics simulations can exhibit marked discrepancies with the real world due to incomplete models, inaccurate parameters, and numerical artifacts \cite{ferguson_intersection-free_2021}; although dynamics randomization can improve sim-to-real transfer, it can require substantial training time and effort \cite{akkaya2019solving, andrychowicz_learning_2020, handa_dextreme_2022} and may penalize precision. Inspired by classical PID control, which can minimize steady-state error and reject disturbances on linear systems, we propose a \textbf{Policy-Level Action Integrator (PLAI)}, which integrates policy actions during an episode.

An established approach for applying policy actions is
\begin{equation}
    s^d_{t+1} = s_{t} \oplus a_{t} = s_{t} \oplus \Pi (o_{t}),
\end{equation}
where $s^d_{t+1}$ is the desired state, $a_t$ is an action expressed as an incremental state target, $s_t$ is the current state, $o_t$ is the current observation, $\Pi$ is the policy, and $\oplus$ computes the state update (e.g., for states defined by position and orientation, $\oplus$ computes composition with a translation and rotation). 

In contrast, \textbf{PLAI} applies policy actions as
\begin{equation}
    s^d_{t+1} = s^d_{t} \oplus a_{t} = s^d_{t} \oplus \Pi (o_{t})
\end{equation}
Thus, the policy action is applied to the \textit{last desired state} instead of the current state. Unrolling from $t=0...T$,
\begin{equation}\label{eq:plai_unrolled}
    s^d_{T} = s^d_0 \oplus \sum_{i=0}^{T-1} a_i = s^d_0 \oplus \sum_{i=0}^{T-1} \Pi(o_i).
\end{equation}
where $s^d_0$ is set to $s_0$\footnote{The summation symbol in Equation~\ref{eq:plai_unrolled} is used as shorthand for successive compositions with actions.}. Thus, the desired state at time $T$ is equal to the initial state composed with successive actions over time, effectively integrating them. (Note that this formulation is \textit{not} open-loop control, as the policy continues to be evaluated on the current observation $o_t$ when generating actions.) When coupled with a low-level PD controller (e.g., a TSI controller), \textbf{PLAI} has a close relationship with a standard (non-integrating) policy coupled with a PID controller; a derivation is provided in Appendix~\ref{sec:plai-derivation}. Empirically, \textbf{PLAI} requires minimal implementation effort (1-2 lines of code), is simple to tune, and outperforms standard PID in our application. %

Like PID controllers, \textbf{PLAI} can experience \textit{windup} when in contact with the environment; unbounded error accumulates, resulting in unstable dynamics.
To mitigate this effect, we also develop \textbf{Leaky PLAI}, which clamps the accumulated control effort. Equations are derived in Appendix~\ref{sec:plai-derivation}. 

\textbf{Evaluation:} From careful observations, discrepancies in simulated and real-world dynamics are primarily caused by nonlinear friction and imperfect gravity compensation on the real robot. Thus, we trained a \textbf{Reach} policy under ideal conditions in simulation, and evaluated the ability of \textbf{PLAI} to reject friction and gravity disturbances in simulation and reality at test time. We evaluated three test-time methods:

\begin{itemize}
    \item \textbf{Nominal:} Actions are applied to the current state.
    \item \textbf{PID:} Actions are applied to the current state. PID is used with classical anti-windup for best-case performance.
    \item \textbf{PLAI:} Actions are applied to the \textit{desired} state.
\end{itemize}

We evaluated each method under three test-time conditions: 
\begin{itemize}
    \item \textbf{Ideal:} No friction or gravity perturbations are applied.
    \item \textbf{Friction:} Joint friction of 0.15 $Nm$ is applied to all robot joints (within the range of identified values from \cite{gaz2019dynamic}).
    \item \textbf{Gravity:} A gravitational perturbation of 0.12 $m/s^2$ is applied to all robot links.
\end{itemize}

\begin{figure}
    \includegraphics[width=\columnwidth]{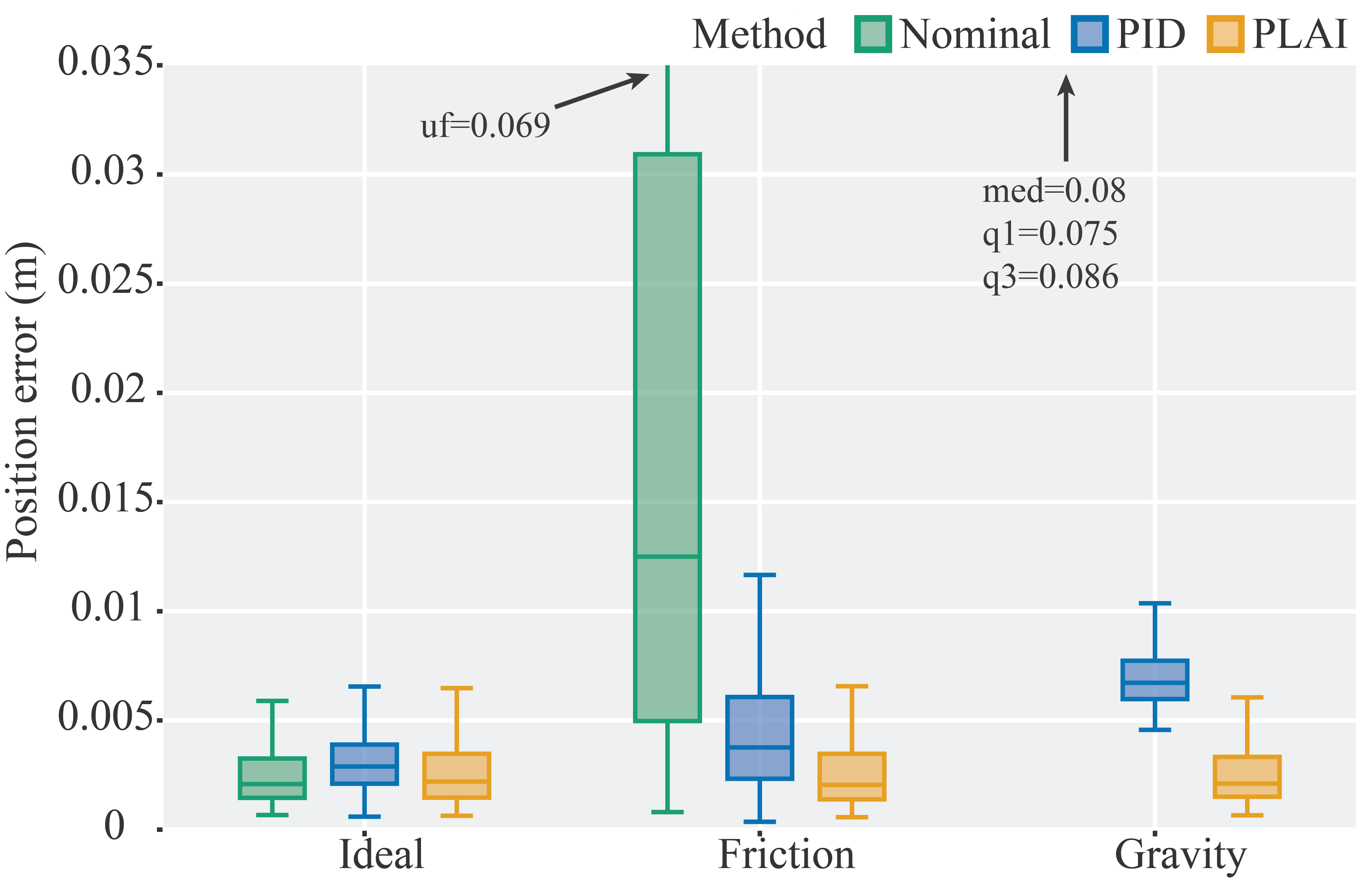}
    \caption{\footnotesize \textbf{Evaluation of PLAI in simulation.} Results of \textbf{Nominal} are annotated when outside of plot bounds. Full-axis plot is in \autoref{fig:control-ablation-full-appendix}.}
    \label{fig:control-ablation}
\end{figure}

We randomized the initial pose and target pose of the robot and measured steady-state position error (\autoref{fig:control-ablation}). Notably, \textbf{PLAI} had substantially lower error and variance than \textbf{Nominal} under friction and gravity, had consistently lower error than PID, and maintained $\approx 2\,mm$ error across all test conditions.

\begin{figure}
    \includegraphics[width=\columnwidth]{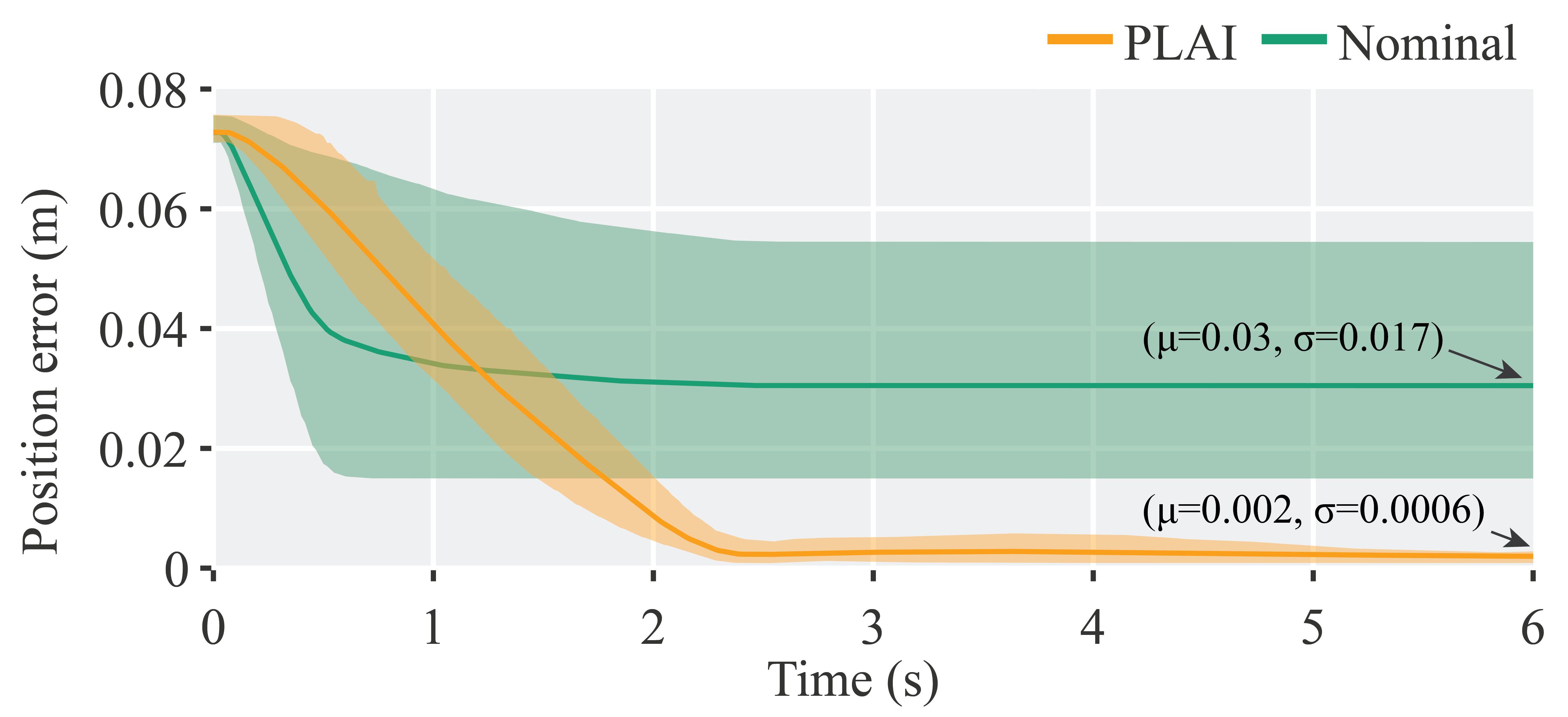}
    \caption{\footnotesize{\textbf{Evaluation of PLAI in the real world.} Each method was tested on 3 different goals with 20 trials each. Evaluation parameters are in \autoref{tab:reach_eval_params}.}}
    \label{fig:control-timeseries-real}
\end{figure}

Next, we conducted a similar experiment in the real world. The \textbf{Reach} policy was deployed with and without \textbf{PLAI} (\autoref{fig:control-timeseries-real}); friction and gravity perturbations were simply from real-world dynamics. Again, \textbf{PLAI} demonstrated substantially lower error and variance than \textbf{Nominal}, with $\approx 2\,mm$ error. As a final comparison, the TSI implementation provided by Franka (RL-free) was deployed on the same task and resulted in $4.45\,mm$ error.
Thus, \textbf{PLAI} is a simple but highly effective means to minimize error with respect to policy targets; moreover, it can be applied exclusively at deployment time.

\section{Real-World Experiments}
\label{sec: eval-insertion-real}

After developing and validating our algorithms, we performed comprehensive experiments and demos to evaluate our real-world system (Figure~\ref{fig:robot_setup}). Five types were executed: \textbf{Pick}, \textbf{Place}, \textbf{Sort}, \textbf{Insert}, and \textbf{Pick-Place-Insert (PPI)}.

\subsection{Pick Experiment}
This experiment evaluated the ability of the real-world system to initiate contact and pick up arbitrarily-placed objects. 

\textbf{Experimental Setup:} 6 different pegs were randomly placed on top of an optical breadboard with dimensions 450 mm x 300 mm, which itself was located within bounds of 500 mm x 350 mm. The pegs were located in trays that were free to slide; angular perturbations of $\pm$10 deg were applied to the trays containing non-axisymmetric pegs. The goal was for the robot to detect all the pegs and use the simulation-trained \textbf{Pick} policy to pick up the objects before releasing them. 

\textbf{Key Results:}
The system demonstrated extremely high success rates (98.8\%) across all pegs (\autoref{tab:eval-real}). Failure cases were one missed detection of a peg, as well as one grasp of both a peg and its corresponding peg tray.

\subsection{Place Experiment}
This experiment evaluated the ability of the real-world system to accurately reach low target locations while maintaining contact with a typically-sized object in the gripper. 

\textbf{Experimental Setup:} Eight 25 mm x 25 mm trays were randomly placed on top of the breadboard. 20 mm x 20 mm printed targets were centered on top of the trays; the targets were used to measure positional accuracy and consisted of concentric rings, each with a thickness of 2 mm. A laser was rigidly mounted to the grippers (\autoref{fig:real_place_eval_data_collection}). The goal was for the robot to detect the trays and use the simulation-trained \textbf{Place} policy to guide the laser to the centers of the targets.

\textbf{Key Results:}
The system demonstrated low steady-state errors, with a mean distance-to-goal of 4.23 $\pm$ 1.96 mm. The error distribution is illustrated in \autoref{fig:eval_place}b.

\subsection{Sort Demonstration}
This experiment qualitatively demonstrated the ability of the robot to execute a realistic sorting procedure.

\textbf{Experimental Setup:} 6 different pegs and 3 different gears were randomly placed on top of the breadboard. The pegs were located in trays, and angular perturbations of $\pm$45 deg were applied. Bins were placed at approximately-determined positions in the workspace. The goal was for the robot to use its \textbf{Pick} and \textbf{Place} policies to detect, pick, place, and drop the round pegs, rectangular pegs, and gears into separate bins.

\textbf{Key Results:} %
Performance was highly repeatable in practice; please see the supplementary video.

\subsection{Insert Experiment}
This experiment evaluated the ability of the real-world system to insert diverse plugs into corresponding sockets, as well as generalize to unseen assets. 

\textbf{Experimental Setup:} 6 different pegs, 3 different gears, and 2 different \textit{unseen} NEMA connectors (2- and 3-prong) were placed imprecisely in the gripper fingers. Holes, gearshafts, and receptacles were mounted to the breadboard. The end-effector was manually guided until the plugs were inserted into their respective sockets; the end-effector pose was recorded as a target. The end-effector was then commanded to a random initial state (\autoref{tab:exp-params-real}).
The robot received an observation of the target with random $X$- and $Y$-axis noise $\sim U[-2, 2]~mm$.\footnote{To our knowledge, only two other sim-to-real efforts have examined perturbations of this magnitude \cite{davchev_residual_2021, zhang_learning_2021}. Both manipulated much larger pegs.} The goal was for the robot to use its \textbf{Insert} policies to insert the plugs into their corresponding sockets.

\textbf{Key Results:} The system demonstrated high engagement rates (i.e., partial insertions) across the pegs (86.7\%), gears (95.0\%), and connectors (100\%), as well as moderately-high success rates (i.e., full insertions) across the same objects (76.7\%, 92.5\%, and 85\%). The \textbf{Pegs and Holes} assembly \textbf{Insert} policy also successfully generalized to NEMA connectors, which can be considered extensions of peg-in-hole.

Engagement failures were almost exclusively due to slip between the gripper and object; we hypothesize that a high-force gripper (e.g., Robotiq) would fully resolve this issue. Full-insertion failures were almost exclusively due to the wedging phenomenon, a longstanding topic of research \cite{whitney1982quasi}.

\begin{figure*}
    \includegraphics[width=\textwidth]{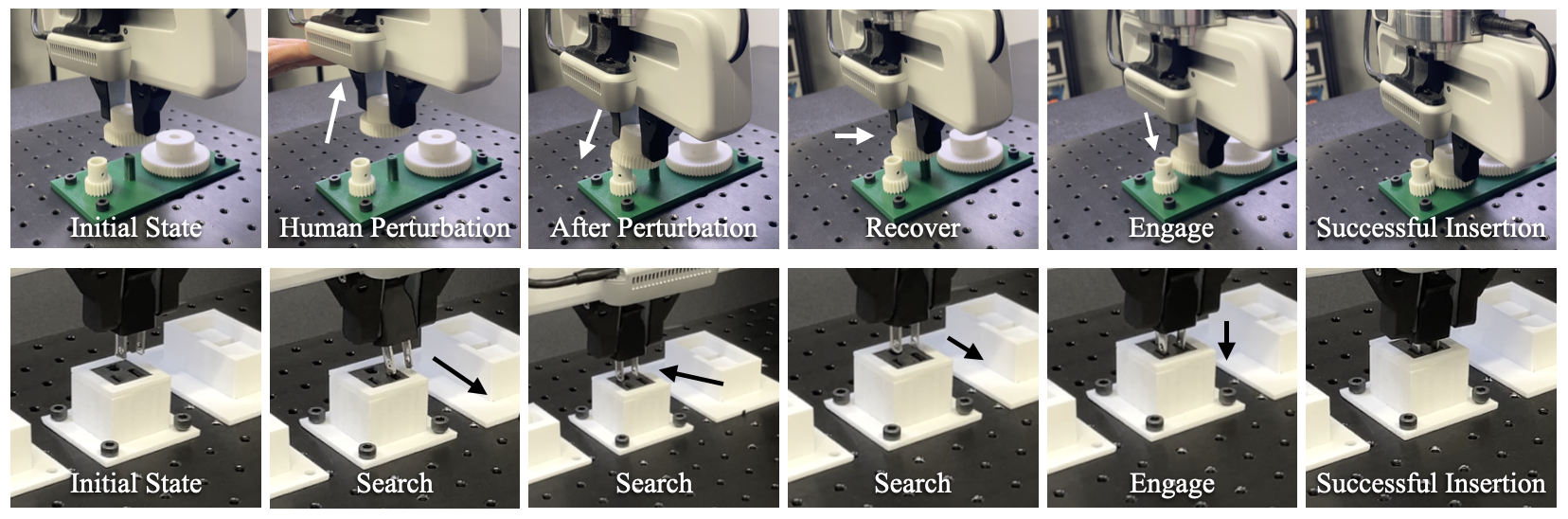}
    \caption{\footnotesize{\textbf{Snapshots of real-world experiments.} Top row: recovery behavior exhibited by the robot after human perturbation during gear insertion. Bottom row: search behavior exhibited during 3-prong NEMA connector insertion.}}
    \label{fig:real_eval_behavior}
\end{figure*}

Informally, when the robot was intentionally perturbed by a human during the \textbf{Gears and Gearshafts} assembly \textbf{Insert} policy, the policy exhibited recovery behavior. In addition, the policy exhibited search behavior on the surface of the socket tray, exploring the vicinity of the observed goal (\autoref{fig:real_eval_behavior}).

\subsection{Pick, Place, and Insert (PPI) Demonstration}
This experiment demonstrated the ability of the robot to execute end-to-end assembly; the \textbf{Insert} policies required robustness to error accumulated from \textbf{Pick} and \textbf{Place}.

\textbf{Experimental Setup:} 6 pegs, 3 gears, 2 NEMA connectors, and their corresponding sockets were initialized with the same conditions as the \textbf{Pick} experiment. The goal was for the robot to bring all parts into their assembled configurations.

\textbf{Key Results:} The system demonstrated even higher success rates than during the \textbf{Insert} experiment: 80\% and 88.3\% success/engagement rates for peg insertion, 97.5\% and 100\% success/engagement rates for gears, and 100\% success/engagement rates for connectors. The higher success rates suggest that the randomization and noise ranges during the \textbf{Insert} experiment may have been particularly adverse.

To our knowledge, \textbf{IndustReal} is the first system to demonstrate RL-based sim-to-real transfer for the end-to-end assembly task (i.e., detection, grasping, part transport, and insertion) without any policy adaptation phase in the real world.

\begin{table}[!htb]
\rowcolors{3}{white}{Gainsboro}
\bgroup
\def\arraystretch{1.14}%
\centering
\scalebox{0.85}{
\begin{tabular}{ l|c|cc|cc } 
\toprule
\multirow{2}{*}{Asset} & \multicolumn{1}{c}{\textbf{Pick}} & \multicolumn{2}{c}{\textbf{Insert}} & \multicolumn{2}{c}{\textbf{Pick-Place-Insert}} \\ 
& Success & Success & Engage & Success & Engage \\ \midrule
Round peg 8 mm & 19/20 & 7/10 & 7/10 & 7/10 & 7/10\\ 
Round peg 12 mm & 19/20 & 7/10 & 9/10 & 7/10 & 7/10 \\ 
Round peg 16 mm & 20/20 & 8/10 & 10/10 & 8/10 & 10/10 \\ 
Rectangular peg 8 mm & 20/20 & 8/10 & 9/10 & 10/10 & 10/10 \\ 
Rectangular peg 12 mm & 20/20 & 8/10 & 8/10 & 8/10 & 9/10 \\ 
Rectangular peg 16 mm & 20/20 & 8/10 & 9/10 & 8/10 & 10/10 \\ 
NEMA 2-prong & - & 10/10 & 10/10 & 10/10 & 10/10 \\ 
NEMA 3-prong & - & 7/10 & 10/10 & 10/10 & 10/10 \\\midrule
Small gear & - & 8/10 & 9/10 & 10/10 & 10/10 \\
Medium gear & - & 9/10 & 9/10 & 9/10 & 10/10 \\
Large gear & - & 10/10 & 10/10 & 10/10 & 10/10 \\ 
Multi-gear assembly & - & 10/10 & 10/10 & 10/10 & 10/10 \\\midrule
Total \# & 158/160 & 100/120 & 110/120 & 107/120 & 113/120 \\ 
\textbf{Total} (\%) & \textbf{98.75\%} & \textbf{83.33\% }& \textbf{91.67\% }& \textbf{89.16\%} & \textbf{94.17\%}\\ 
\bottomrule
\end{tabular}}
\egroup
\caption{\footnotesize{\textbf{Real-world experimental results for Pick, Insert, and PPI.}}}
\label{tab:eval-real}

\hfill

\end{table}

\section{IndustRealKit and IndustRealLib}
\label{sec:industreal}

We strongly encourage fellow researchers to reproduce our results and use our platform to investigate their own questions. As follows, we will open-source our two most critical pieces of hardware and software, \textbf{IndustRealKit} and \textbf{IndustRealLib}.

\textbf{IndustRealKit} contains 3D-printable CAD models for all parts we designed, as well as a list of all parts we purchased from external vendors (Figure~\ref{fig:factory_kit}). The CAD models include 20 parts: 6 peg holders, 6 peg sockets (i.e., extruded holes), 3 gears, 1 gear base (with gearshafts), and 4 NEMA connectors and receptacle holders. 
The purchasing list includes 17 parts: 6 metal pegs (from the NIST benchmark), 4 NEMA connectors and receptacles, 1 optical breadboard, and fasteners.

\textbf{IndustRealLib} is a lightweight library containing code for policy deployment and training. Specifically, we provide scripts to allow users to quickly deploy policies from Isaac Gym \cite{makoviychuk_isaac_2021} onto a Franka robot.
The scripts include a base class that implements our policy-level controllers and sends/receives actions and observations from \textit{FrankaPy}; task-specific classes that interpret actions from the policy, compute observations, and set targets; and a script that instantiates the classes, loads corresponding policies, and executes them. For the \textbf{Peg and Hole} assemblies, we also provide weights for the \textbf{Reach}, \textbf{Pick}, \textbf{Place}, and \textbf{Insert} policies. We have thus far used \textbf{IndustRealLib} on two different Franka robots in two different cities. %
Finally, we provide code for training our RL policies, including implementations of \textbf{SAPU}, \textbf{SDF-Based Reward}, and \textbf{SBC}. This code also includes a carefully-reviewed Franka model and simulation parameters validated during this work.

\section{Limitations \& Future Work}
\label{sec:limitation}
Our work has limitations, which lend themselves naturally to future research directions. First, like other efforts on sim-to-real transfer for assembly tasks, we have primarily investigated tasks inspired by the NIST benchmark \cite{kimble_benchmarking_2020}. However, recent work in graphics and robotics has provided a large number of simulation-compatible assets for assembly (e.g., \cite{tian2022assemble}), potentially enabling RL policies that can generalize across widely different categories. Second, our primary failure cases on the real system were due to slip of the object in the gripper and wedging of plugs in their corresponding sockets. We believe that these cases can be resolved by providing the agent with simulated visuotactile readings during training \cite{si2022taxim, xu2022efficient, wang2022tacto} and using corresponding sensors on the real-world system \cite{yuan2017gelsight, lambeta2020digit}, as well as more accurately simulating friction \cite{andrews2022contact}. Third, we do not explore passive mechanical compliance as a means for facilitating policy learning \cite{morgan2021vision};
we believe that optimizing the policy, controller, and passive dynamics simultaneously
can significantly help improve task performance. Fourth, our sim-to-real framework currently relies on a high-accuracy simulator and our proposed training- and deployment-time algorithms. However, for some tasks of even higher complexity (e.g., assembly of elastic cables onto pulleys), the simulator may neither be fundamentally accurate nor efficient enough to smoothly train and deploy policies to the real world. We envision the construction of a tight feedback loop from real-world deployments (i.e., a sim-to-real-to-sim loop) as a potentially compelling training strategy \cite{abeyruwan2022sim2real, lim2022real2sim2real}.

\section{Conclusions}
\label{sec:conclusions}
In this paper, we have presented \textbf{IndustReal}, a set of algorithms, systems, and tools to solve benchmark assembly problems in simulation and transfer policies to the real world. The utility of our simulation-based algorithms (\textbf{SAPU}, \textbf{SDF-Based Dense Reward}, and \textbf{SBC}) and real-world algorithm (\textbf{PLAI}) has been demonstrated through careful experiments in simulation and the real world. We provide the first simulation results for a series of benchmark tasks proposed in \cite{narang2022factory}, and most critically, we demonstrate what is, to our knowledge, the first real-world system for RL-based sim-to-real on the end-to-end assembly task with no policy adaptation phase. Finally, in the hope of full reproducibility, we provide hardware and software for others in the community to replicate our results.

\section*{Acknowledgments}

The authors thank Michael Noseworthy for advice throughout the project, Bowen Wen and Ajay Mandlekar for advice on perception, Eric Heiden and Balakumar Sundaralingam for advice on writing Warp kernels for SAPU, Lucas Manuelli for help with the initial implementation of IndustRealLib and advice on dynamics and control, Karl Van Wyk and Nathan Ratliff for feedback on PLAI, Viktor Makoviychuk for advice on domain randomization and asymmetric actor-critic, Gavriel State and Kelly Guo for providing support with Isaac Gym, Philipp Reist and Tobias Widmer for providing support with PhysX, and Sandeep Desai and Kenneth Maclean for providing support with hardware setup and 3D printing.

\section*{Contributions}
Bingjie T., Yashraj N., Fabio R. developed SAPU.

Bingjie T., Yashraj N., Dieter F. developed SDF reward.

Bingjie T., Yashraj N. developed SBC.

Michael L., Yashraj N. developed PLAI.

Michael L., Yashraj N., Iretiayo A., Bingjie T. developed IndustRealLib.

Yashraj N., Michael L. developed IndustRealKit.

Bingjie T., Yashraj N., Ankur H. developed the perception module.

Bingjie T., Michael L. ran experimental evaluations.

Yashraj N., Bingjie T., Michael L. wrote the paper.

Yashraj N., Bingjie T., Michael L., Fabio R., Ankur H., Gaurav S. revised the paper.

Bingjie T., Yashraj N., Michael L. created the video.

Yashraj N., Dieter F., Fabio R., Gaurav S., Ankur H., Iretiayo A. advised the project.

\bibliographystyle{plainnat}
\bibliography{references}
\pagebreak
\appendix
\renewcommand\thefigure{S\arabic{figure}}

\subsection{Related Work}
\label{sec:appendix_related_work}

Here we review classical and learning-based approaches to robotic assembly and briefly comment on simulation.

\subsubsection{Classical Approaches}

Assembly has been an open challenge in robotics for decades \cite{whitney_mechanical_2004, mason2001mechanics}. Many analytical methods have been proposed to solve assembly tasks, particularly for the canonical problem of peg-in-hole insertion; these methods have typically utilized geometry, dynamics, mechanical design, and sensing as fundamental tools. \citet{drake1978using} proposed remote center compliance (RCC) as a means to mitigate reliance on visual or force sensing during assembly. \citet{whitney1982quasi} described the effects of part geometry, gripper and support stiffness, and friction on contact forces and adverse outcomes (e.g., jamming and wedging). \citet{lozano1984automatic} proposed compliant motion planning strategies for assembly. \citet{xia2006dynamic} derived a compliant contact model and used the model to avoid jamming and wedging. \citet{huang2013fast} addressed initial part misalignment using vision, and \citet{tang2016autonomous} addressed this challenge via force/torque sensing. The preceding principles and methods are the prevailing means of addressing the annual NIST Assembly Task Board challenge, the established benchmark in robotic assembly \cite{kimble_benchmarking_2020, von2020robots}. Nevertheless, such methods can be highly sensitive to errors in modeling, sensing, and state estimation; perturbations of adapters, fixtures, and calibration; and the introduction of unseen or more complex assets.

\subsubsection{Learning-Based Approaches}

In the past few years, learning-based approaches to assembly have gained popularity in the robotics research community, with many of the efforts focused on RL. Earlier works have typically explored model-based algorithms, such as guided policy search (GPS) \cite{thomas_learning_2018} and iterative linear-quadratic-Gaussian control (iLQG) \cite{luo_reinforcement_2019}. Despite the sample efficiency of these algorithms, nonlinear and discontinuous contact dynamics have made them challenging to leverage for contact-rich manipulation tasks \cite{spector_deep_2020}.

A number of recent works have used model-free, off-policy RL algorithms, including classical Q-learning \cite{inoue_deep_2017}, deep-Q networks (DQN) \cite{zhang_learning_2021}, soft actor-critic (SAC) \cite{beltran-hernandez_variable_2020}, probabilistic embeddings for actor-critic RL (PEARL) \cite{schoettler2020meta}, hierarchical RL \cite{hou_data-efficient_2021}, and most popularly, deep deterministic policy gradients (DDPG) \cite{apolinarska_robotic_2021, luo_learning_2021, luo2021robust, vecerik_practical_2019}. These algorithms are also sample efficient, but can have unfavorable convergence properties.

A smaller number of research efforts have used model-free, on-policy algorithms, including trust region policy optimization (TRPO) \cite{lee_making_2020}, proximal policy optimization (PPO) \cite{hebecker_towards_2021, son_sim-to-real_2020, vuong_learning_2021, narang2022factory}, and asynchronous advantage actor-critic (A3C) \cite{shao_learning_2020}. These algorithms have favorable convergence properties and are easy to tune; however, they are highly sample inefficient and can require long training times.

Other recent works have used on-policy or off-policy RL algorithms that can leverage demonstrations (e.g., human demonstrations or reference trajectories and controllers), such as residual learning from demonstration (rLfD), minimum-jerk trajectories, or impedance controllers, \cite{davchev_residual_2021, kozlovsky2022reinforcement, johannink2019residual}, interleaved Riemannian Motion Policies (RMP) and SAC \cite{lee2020guided}, guided DDPG \cite{fan_learning_2019}, DDPG from demonstration (DDPGfD) \cite{luo2021robust, luo_dynamic_2019, vecerik_leveraging_2018}, offline meta-RL with advantage-weighted actor-critic (AWAC) \cite{zhao2022offline}, and inverse RL \cite{wu_learning_2021}. For efforts using human demonstrations, substantial engineering infrastructure and collection time can be required; demonstrations can be suboptimal; and successful demonstrations can be difficult to reliably obtain during high-precision tasks.

Several learning-based, non-RL approaches have also been proposed. These approaches include using human-initialized self-supervised learning to learn a policy or residual policy with multimodal inputs \cite{spector2021insertionnet, spector2022insertionnet, fu2022safely}, as well as learning from videos of human demonstrations using category-level visual representations and 6-DOF tracking \cite{wen2022you}.

The preceding efforts define the state-of-the-art in learning-based approaches for assembly. Several have demonstrated high success rates and repeatability, shown robustness to small perturbations of initial part poses, and/or shown some degree of generalization across parts; one has even outperformed a solution from professional integrators \cite{luo2021robust}. However, most successful efforts have required human initializations, demonstrations, or on-policy corrections. Furthermore, purely real-world approaches are inherently difficult to parallelize; may require long training to achieve appreciable robustness (e.g., $\sim$50 hours in \cite{luo2021robust}); typically require manual resets; can be impractical for time-consuming, expensive, delicate, or dangerous tasks (e.g., construction); and do not fully leverage the substantial amount of virtual data available for industrial settings (e.g., nearly every existing industrial part originates from a CAD model that can be rendered or simulated).

\subsubsection{Simulation}

To our knowledge, the state-of-the-art in accurate and efficient contact-rich simulation is captured in \cite{narang2022factory, macklin_local_2020, chen2022midas, lan2022affine, tian2022assemble, yoon2022fast}. Among these, \cite{narang2022factory, chen2022midas} specifically address simulation for robotics tasks, whereas \cite{narang2022factory} integrates these capabilities within a widely-used robotics simulator \cite{makoviychuk_isaac_2021}. Thus, we leverage \cite{narang2022factory} as our simulation platform.

\subsection{Real-World System}

\subsubsection{Communications}

\begin{figure}[H]
\centering
    \includegraphics[width=.45\textwidth]{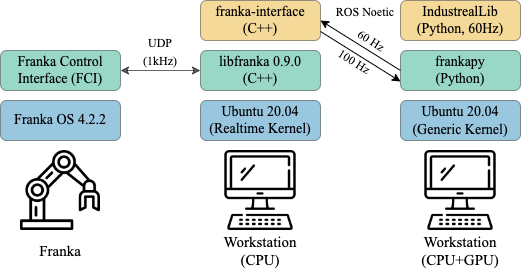}
\caption{Schematic of communications framework.}
\label{fig:communications}
\end{figure}

A schematic of our communications framework is shown in Figure~\ref{fig:communications}. The input to the communications pipeline is a trained RL policy (specifically, a checkpoint file) from Isaac Gym, which is provided to IndustRealLib. The output of the communications pipeline is a set of torque commands communicated to the robot.

\begin{figure}[H]
\centering
     \begin{subfigure}[b]{0.24\textwidth}
     \centering
    \includegraphics[width=0.99\textwidth]{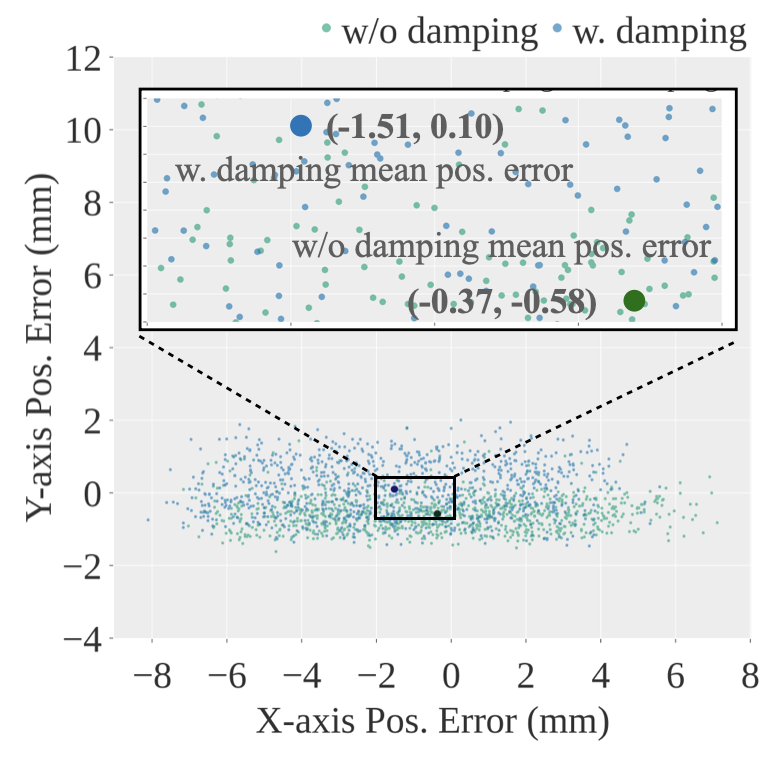}
    \caption{Simulation (1000 trials)}
    \label{fig:sim_eval_place}
     \end{subfigure}
     \begin{subfigure}[b]{0.24\textwidth}
     \centering
    \includegraphics[width=0.99\textwidth]{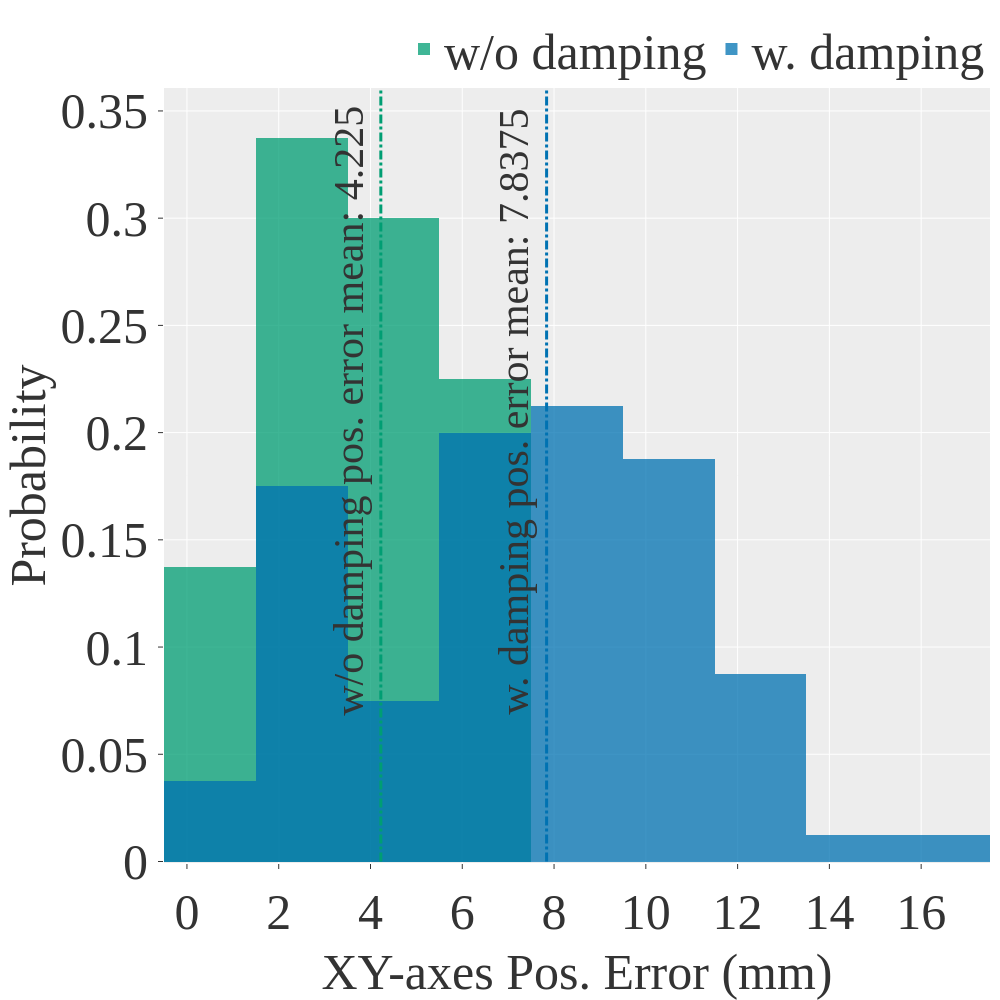}
    \caption{Real world (160 trials)}
    \label{fig:real_eval_place}
     \end{subfigure}
    \caption{Evaluation of \textbf{Place} policy in simulation and the real world, with/without damping during simulation-based training.}
    \label{fig:eval_place}
\end{figure}

\subsubsection{Camera Calibration}
\label{appendix:camera-calibration}

The goal of intrinsic camera calibration for an RGB camera is to determine the relationship between the location of a 3D point in space and its location in the image. We rely on the intrinsic camera parameters provided by Intel through the \textit{librealsense} and \textit{pyrealsense2} libraries.

The goal of extrinsic camera calibration is to determine the 6-DOF pose $T_{cam}^{robot}$ of the camera with respect to the robot frame. To compute extrinsic camera parameters, we first place a 6-inch AprilTag (52h13) onto the work surface and command the robot via the \textit{frankx} library to random 6-DOF poses in the robot workspace, biased towards having the tag in view, but avoiding direct overhead views due to pose ambiguity. At each viewpoint, we detect the tag and compute the 6-DOF pose $T_{tag}^{cam}$ of the tag with respect to the camera frame via the \textit{pupil-apriltags} library, and we simultaneously query the 6-DOF pose $T_{ee}^{robot}$ of the end-effector in the robot base frame. We collect approximately 30 such samples and then use the Tsai-Lenz method \cite{tsai1989new} from the \textit{OpenCV} library to compute the 6-DOF pose $T_{cam}^{ee}$ of the camera with respect to the end-effector. The 6-DOF pose $T_{cam}^{robot}$ of the camera with respect to the robot base frame can then simply be computed as $T_{ee}^{robot} T_{cam}^{ee}$.

We validated the extrinsic parameters by comparing our estimated $T_{cam}^{ee}$ against the corresponding transformation measured in a CAD assembly containing part models of the RealSense camera, camera mount, and end-effector. 

\subsubsection{Object Detection}

The goal of our object detection module is to determine the object identities, 2D bounding boxes, and segmentation masks of each of our parts in the RGB image. To perform object detection, we used the implementation of the well-established Mask R-CNN \cite{he2017mask} network architecture available in the \textit{torchvision} library, which was pretrained on the Microsoft COCO dataset \cite{lin2014microsoft}.

As the COCO dataset does not contain industrial assets, initial tests on our parts resulted in failure. Thus, we fine-tuned the pretrained model on real-world images of our assets. Specifically, we used the RealSense to capture 10-30 overhead images of each part randomly placed on our work surface, as well as 10 images of the work surface itself. We used Adobe Photoshop to automatically remove the backgrounds from the part images and extracted bounding boxes from the results.

We then divided the images into three different sets: 1) {background, round pegs, rectangular pegs, round holes, and rectangular holes}, 2) {background, small gear, medium-sized gear, and large gear}, and 3) {background, NEMA 1-15 plug, NEMA 1-15 socket, USB-C plug, and USB-C socket}.

For each set, we generated an augmented collection of images that consisted of each non-background element with random translations, rotations, and scaling. The elements were overlaid upon randomly-selected background images, and the composites were subject to color jitter using the \textit{kornia} library.

Each augmented set of images was then used to train a Mask R-CNN model in \textit{pytorch}. The categories within the set were added to the pretrained model, and the model was fine-tuned on images from the set to minimize losses over object identities, bounding boxes, and segmentation masks. Each model was trained for 50 epochs with 4000 training images, requiring $\approx$ 7.5 hours on a single GPU, at which point precision and recall scores were typically above 85\%. 

When using our trained models at test time, we performed additional data augmentation consisting exclusively of color jitter on captured images. The augmented images were used as input to our detection model. For each object in the image, the image with the highest object-identification confidence score was used to extract the object's identity, bounding box, and segmentation mask. This test-time augmentation improved our robustness to lighting variation and the presence of distractors. The yaw angle of each object was extracted by computing a minimum-area rectangle on the bounding box with \textit{OpenCV} and calculating the angle of the box with the horizontal.

\subsubsection{Workspace Mapping}
\label{appendix:workspace-mapping}

For each detected object, we computed the centroid of its bounding box in image space. As mentioned earlier, the primary goal of our perception pipeline is to estimate the 2D poses ($x$, $y$, $\theta$) of the parts on the workbench in the robot frame; thus, we needed to convert the location of the centroid (as well as the yaw angle determined earlier) from image space to 3D space.

In order to perform this transformation, we placed a 3-inch AprilTag (53h13) at an arbitrary location in the field of view and computed the pose $T_{tag}^{cam}$ of the tag in the camera frame. With additional knowledge of the size of the tag and the pixel resolution of the camera, distances in image space were mapped to distances in the camera frame. (Strictly speaking, this relation holds only within the plane containing the AprilTag, with decreasing accuracy farther away due to perspective transformations.) The pose $T_{tag}^{cam}$, the location of the centroid of a particular part, and the distance mapping were used to convert the centroid and yaw angle from image space to the camera frame. Finally, the location of the centroid and the yaw angle were converted from the camera frame to the robot frame using the transformation $T_{cam}^{robot}$ computed earlier.

We note that our workspace mapping process was susceptible to error induced by yaw rotations of the camera with respect to the robot frame, which resulted in systematic bias of our perceived locations relative to their actual locations. Thus, we performed a final one-time calibration step, during which we executed the Place experiment (i.e., laser tests) near the four corners of the workspace, estimated the offsets relative to the centers of the targets, averaged the offsets, and subtracted the average from all subsequently-computed 3D locations.

\subsubsection{Constructing Targets}
\label{appendix:constructing-targets}

After execution of the perception pipeline, the robot receives corresponding ($x$, $y$, $z=h_n$, $\alpha=\theta$, $\beta=0.0$, $\gamma=0.0$) as end-effector targets.  The nominal height $h_n$ at which to pick or place each part is specified in advance by the human; in our experience, for industrial-style parts, this specification is trivial (e.g., 25-50\% from the top of the part) and requires little to no iteration. 

\subsubsection{Experimental Setup}

\begin{figure}[ht]
    \includegraphics[width=0.9\columnwidth]{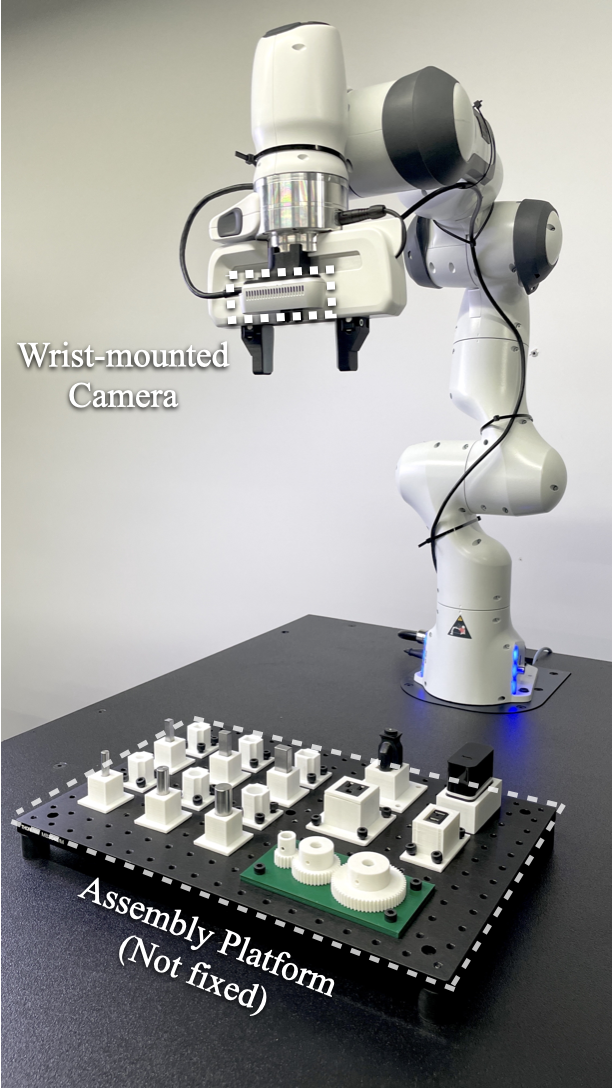}
    \caption{Real-world experimental setup.}
    \label{fig:robot_setup}
\end{figure}

\begin{figure}[ht]
\centering
    \includegraphics[width=0.9\columnwidth]{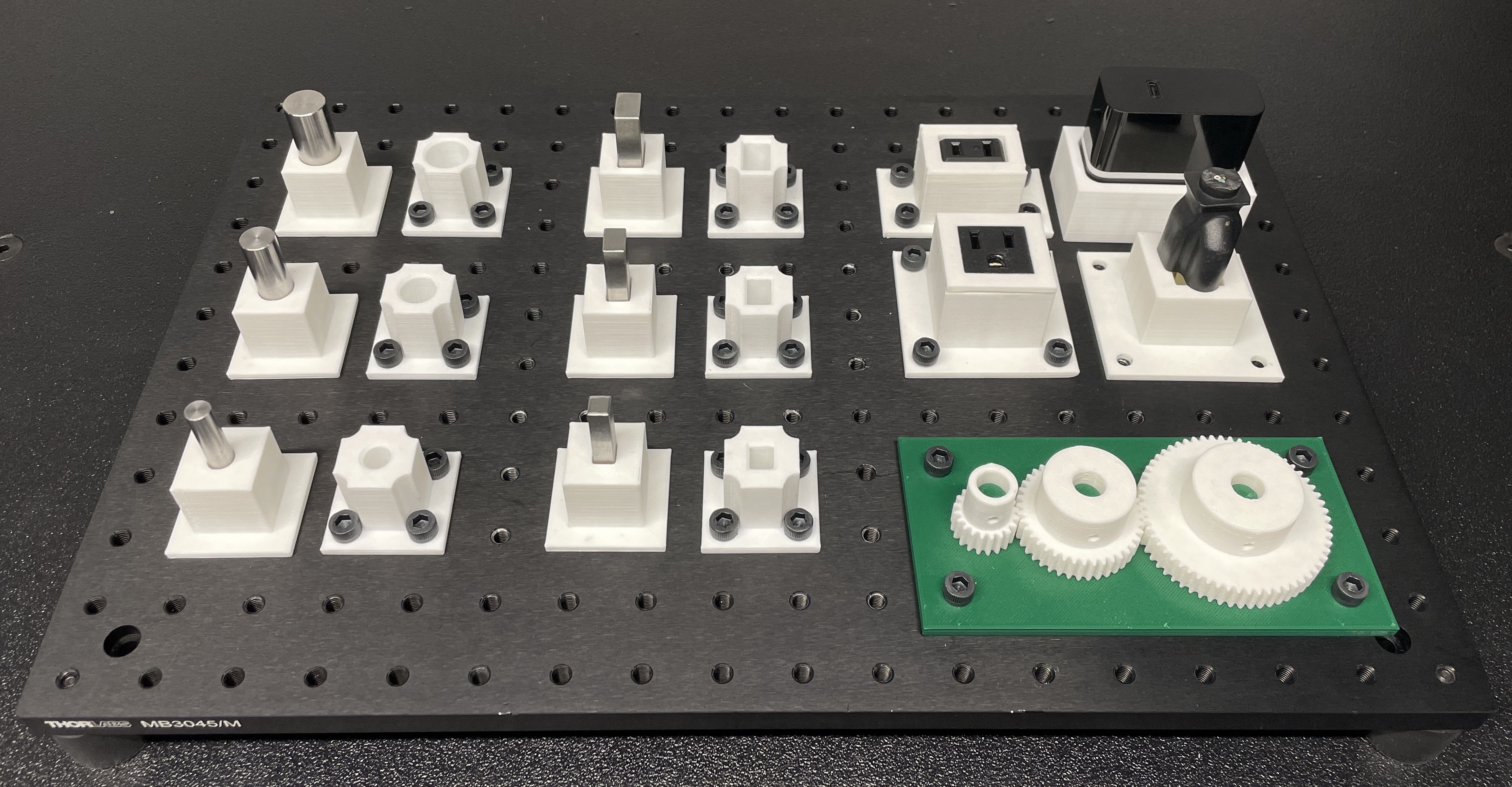}
    \caption{A subset of parts from IndustRealKit.}
    \label{fig:factory_kit}
\end{figure}

Our real-world experimental setup consists of a Franka Emika
Panda arm with a wrist-mounted RGB-D camera, as well as an optical breadboard upon which mechanical parts are placed. The robot, camera, optical breadboard, and parts are shown in \autoref{fig:factory_kit} and \autoref{fig:robot_setup}.

\begin{figure}[ht]
\centering
\includegraphics[width=0.9\columnwidth]{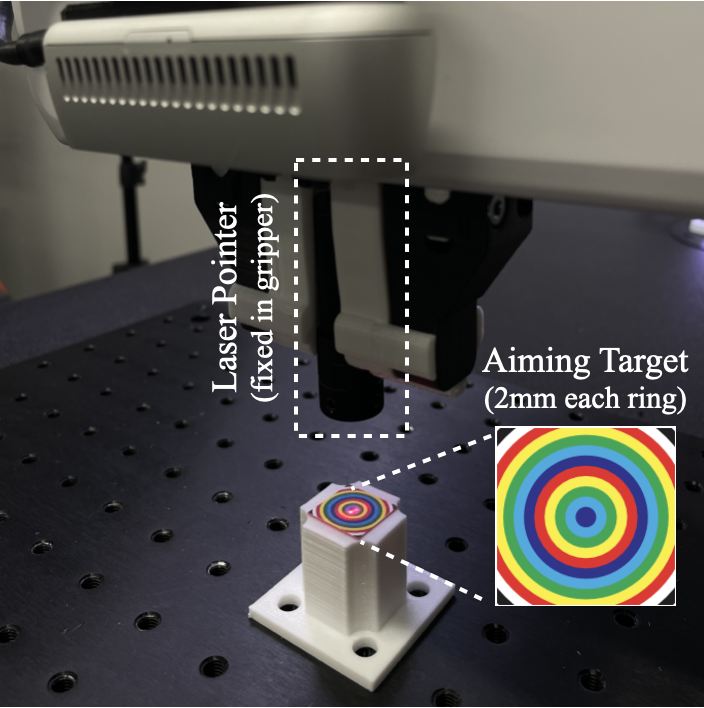}
    \caption{Experimental setup for evaluating \textbf{Place} policy.}
\label{fig:real_place_eval_data_collection}
\end{figure}

\subsubsection{Summary}

We have thus described our full perception pipeline. Camera calibration was performed once before beginning all experiments. The detection and workspace mapping process (aside from the one-time calibration step) were performed at the beginning of each trial that required detections; in total, these per-trial steps took approximately 10 seconds to execute on our non-realtime system.

\subsection{Policy-Level Action Integrator}
\label{sec:plai-derivation}

\subsubsection{Standard PLAI}

Consider an RL policy that generates relative pose actions; in other words, the action composed with the current pose produces the desired state. In practice, the ability to achieve these desired states depends on passive and active system dynamics; having a large discrepancy between the environment used for training and deployment can lead to poor policy performance,
especially when the system uses a more dynamic controller (e.g., a low stiffness task-space impedance controller). In this context, we briefly compare the behavior of \textbf{PLAI} with a lower-level proportional (P) controller, against a standard proportional-integral (\textbf{PI}) controller.\footnote{The comparison would also hold for PLAI with a lower-level PD controller, against a PID controller; we omit derivative terms for simplicity.}

The general form of a discrete-time P controller is given by
\begin{align}
    F[n] &= k_P e[n]
\end{align}
where $F$ is the control effort (e.g., force or torque), $k_P$ is the proportional gain, and $e$ is an error signal.

We define the error signal as the difference between the desired state and the current state:
\begin{align}
    e[n] = x^d[n] - x[n]
\end{align}
where $x$ is the state and superscript $d$ denotes \textit{desired}.

For \textbf{PLAI}, we apply actions to the previous desired state:
\begin{align}
    x^d[n] &= \Pi(o[n])+x^d[n-1]
\end{align}

In other words, we interpret actions as the difference between the current desired state and the previous desired state:
\begin{align}
    \Pi(o[n]) = x^d[n] - x^d[n-1]
\end{align}

We can re-write the error signal as
\begin{align}
    e[n] = \Pi(o[n]) + x^d[n-1] - x[n]
\end{align}

Next, we can roll out the first few timesteps of $F$:
\begin{align*}
    \begin{split}
    F[1] &= k_P\left(\Pi(o[1]) + x^d[0] - x[1]\right)\\
            &= k_P\left(\Pi(o[1]) + \Pi(o[0]) + x^d[0] - x[1]\right)\\
    \end{split}\\
    \begin{split}
    F[2] &= k_P\left(\Pi(o[2]) + x^d[1] - x[2]\right) \\
            &= k_P\left(\Pi(o[2]) + \Pi(o[1]) + \Pi(o[0]) + x^d[0] - x[2]\right)\\
    \end{split}\\
    \begin{split}
    F[3] &= k_P\left(\Pi(o[3]) + x^d[2] - x[3]\right)\\
            &= k_P\left(\Pi(o[3]) + \Pi(o[2]) + ... + \Pi(o[0]) + x^d[0] - x[3]\right)\\
    \end{split}\\
    \begin{split}
    F[N] &= k_P(\Pi(o[N]) + \Pi(o[N-1]) + ... \\
    & \qquad+ \Pi(o[0]) + x^d[0] - x[N])\\
    \end{split}
\end{align*}

More generally,
\begin{equation}
    F[N] = k_P \left(\sum_{k=0}^{N}(\Pi(o[k])) + x^d[0] - x[N] \right)
\end{equation}
In words, the sum of the policy outputs determines the control setpoint, which is tracked by the P controller. The summation term closely resembles an integral term in a \textbf{PI} controller:
\begin{align}
    F[N] &= k_P e[N] + k_I \sum_{k=0}^{N}(e[k])\\
    &= k_P (x^d[N] - x[N]) + k_I \sum_{k=0}^{N}(\Pi(o[k]))
\end{align}

The main difference is that the integral term in \textbf{PLAI} is used as a control setpoint, whereas in \textbf{PI}, it is directly converted into control effort. More extensively,
\begin{enumerate}
\item \textbf{PLAI} integrates the policy actions to generate a setpoint, which is used by the low-level impedance controller to attract the real state towards the desired state. If the system is disturbed from its current state (e.g., the robot is pushed away), the setpoint will not change instantaneously. Instead, the force vector will pull the system towards the same setpoint.
\item \textbf{PI} integrates the policy actions (in this case, equal to the control error) to generate a corrective force vector. If the system is disturbed from its current state, the accumulated error will be applied to an unintended state and may become a disturbance to the policy.
\end{enumerate}

\subsubsection{PLAI for 6-DOF Pose}
The \textbf{PLAI} derivation above applies directly to control of Cartesian position; executing \textbf{PLAI} on Cartesian orientation is very similar, as addition and subtraction operators can be replaced by rotation matrix operations. Specifically, addition can be replaced with
\begin{equation}
    R^O_{e_{n+1}} = R^O_{e_{n}}R^{e_n}_{e_{n+1}} \label{eq:rot-addition}
\end{equation}
and subtraction can be replaced with
\begin{equation}
    R^O_{e_{n}} = R^O_{e_{n+1}}[ R^{e_n}_{e_{n+1}} ]^T \label{eq:rot-subtraction}
\end{equation}
where $R^O_B$ is the rotation of a frame $B$ relative to the frame $O$, and $R^{e_n}_{e_{n+1}}$ is an incremental rotation of the end-effector at time step $n+1$ relative to time step $n$. %

\subsubsection{Leaky PLAI}

After the \textbf{PLAI} update (Equation 3), we can simply rewrite $s_{t+1}^d$ as
\begin{equation}
    \begin{aligned}
        s^d_{t+1} &= s_t \oplus (s^d_{t+1} \ominus s_t) 
    \end{aligned}
\end{equation}
and update the desired state as
\begin{equation}
    \begin{aligned}
        s^d_{t+1} \leftarrow s_t \oplus \min \left( (s^d_{t+1} \ominus s_t), \epsilon \right)
    \end{aligned}
\end{equation}
where $\epsilon$ is a threshold transformation.

\begin{figure}[H]
\centering
    \includegraphics[width=\columnwidth]{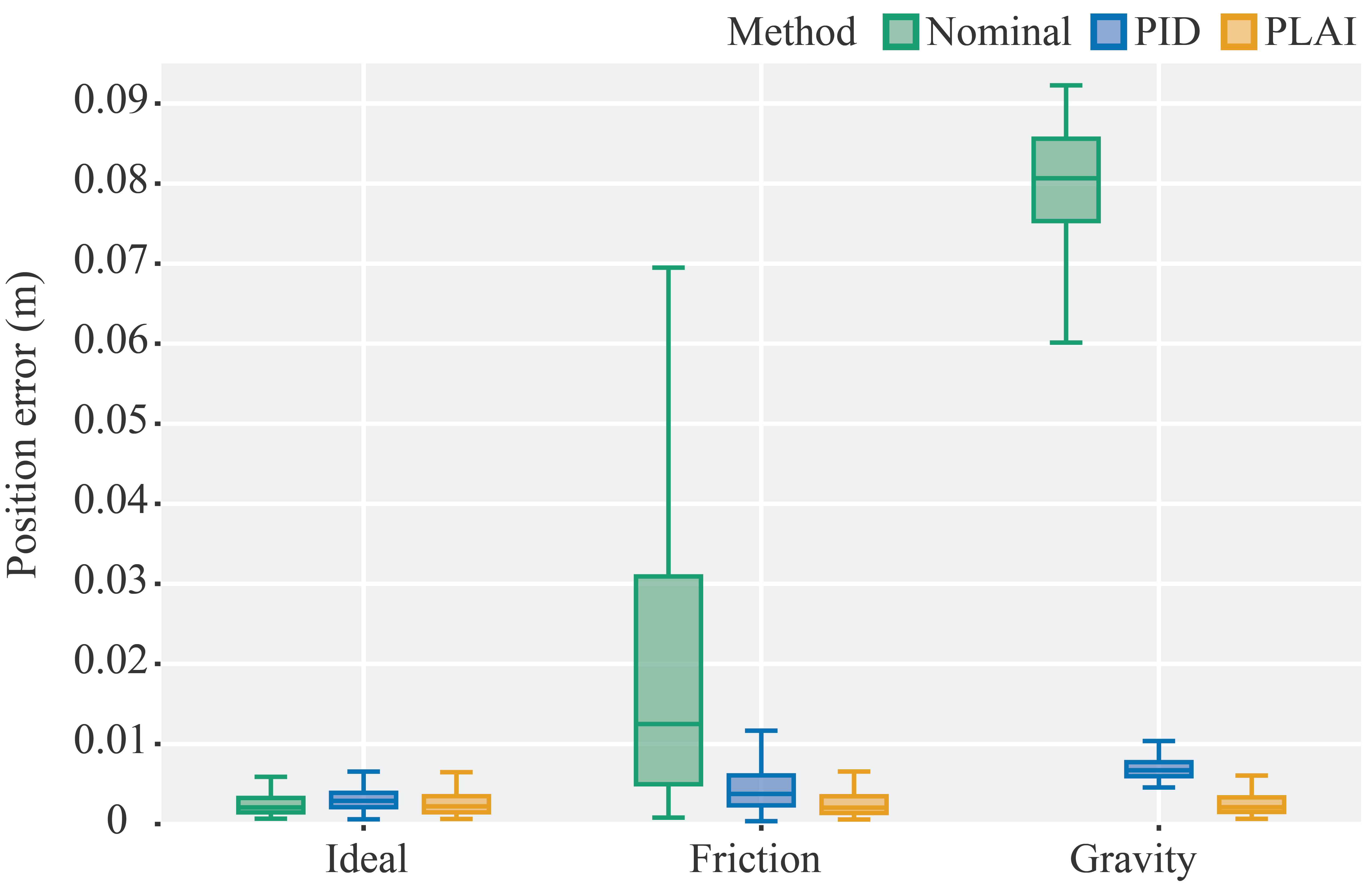}
    \caption{Evaluation of \textbf{PLAI} in simulation. \textbf{PLAI} is compared to \textbf{Nominal} and \textbf{PID} under three environmental conditions. \textit{Ideal} indicates no perturbations.}
    \label{fig:control-ablation-full-appendix}
\end{figure}

\begin{figure}[H]
    \includegraphics[width=\columnwidth]{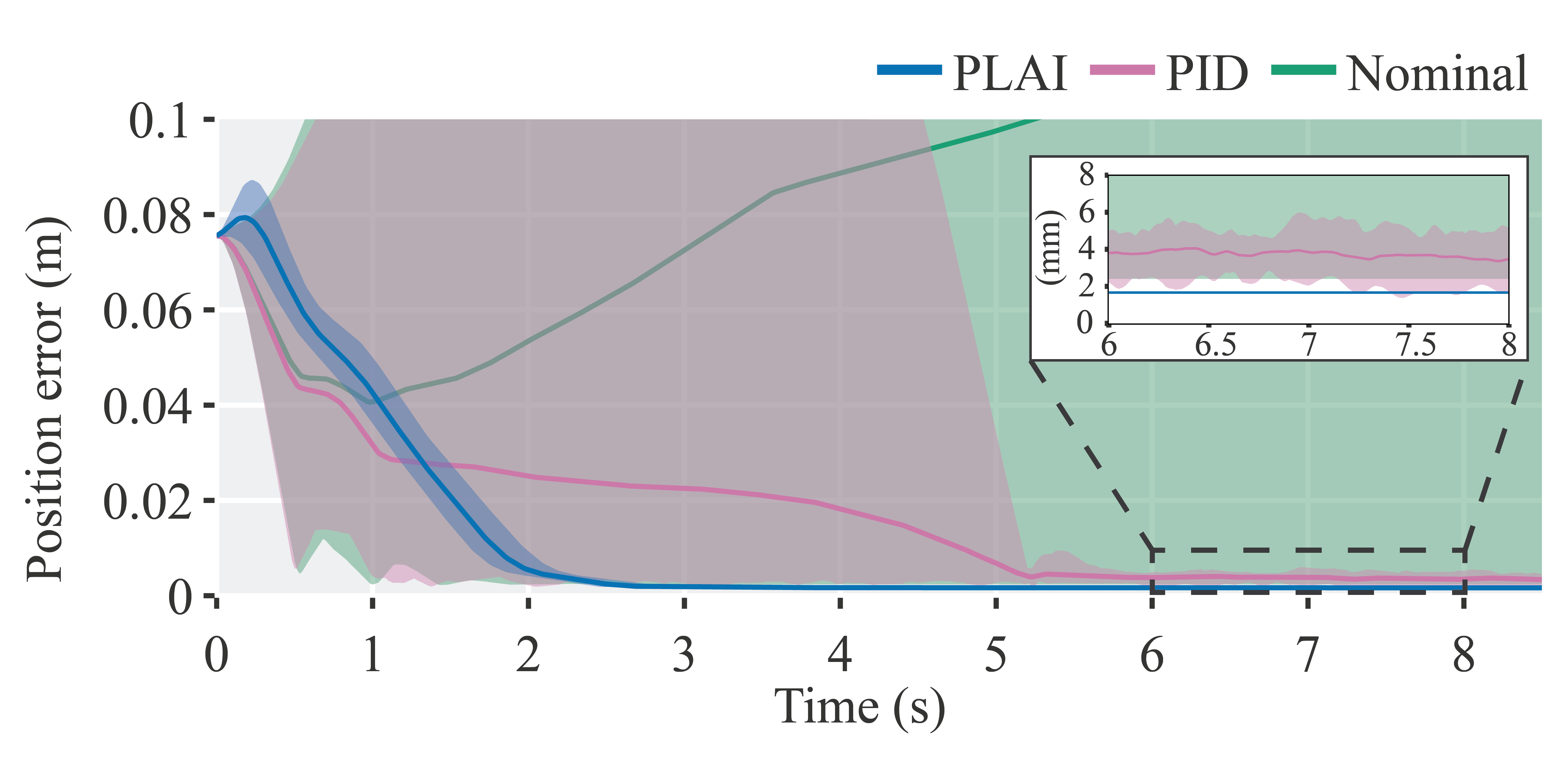}
    \caption{Evaluation of \textbf{PLAI} in simulation, illustrating time-series behavior. \textbf{PLAI} is compared to \textbf{Nominal} and \textbf{PID} over 20 trials under a randomized gravitational disturbance (i.e., gravitational acceleration $g \sim U[0.1, 2.0] m/s^2$). Position error is measured with respect to final target. Inset view shows that steady-state error is lowest for \textbf{PLAI} (1.6 mm), followed by \textbf{PID} (3.5 mm); \textbf{Nominal} is frequently unable to converge.}
    \label{fig:control-timeseries}
\end{figure}

\subsection{Additional Simulation Parameters \& Results}

\subsubsection{MDP formulation \& Parameters in Simulation}
Given our choice of RL, we can formulate the problem as a Markov decision process (MDP) with state space $\mathcal{S}$, observation space $\mathcal{O}$, action space $\mathcal{A}$, state transition dynamics $\mathcal{T}: \mathcal{S} \times \mathcal{A} \rightarrow \mathcal{S}$, initial state distribution $\rho_0$, reward function $r: \mathcal{S}  \rightarrow \mathbb{R}$, horizon length $T$, and discount factor $\gamma \in (0, 1]$. The objective is to learn a policy $\pi : \mathcal{O} \rightarrow \mathbb{P}(\mathcal{A})$ that maximizes the expected sum of discounted rewards $\mathbb{E}_{\pi}[\Sigma^{T-1}_{t=0}\gamma^t r(s_t)]$.

We do not assume that state is fully observable in either simulation or the real world (specifically, $\mathcal{O} \subsetneq \mathcal{S} $). In the real world, the Franka's joint velocities, end-effector velocities, joint torques, and end-effector forces exhibit appreciable noise in free space. In addition, perceptual error leads to noise in object pose estimates, which in turn introduce noise into target poses. Furthermore, without tactile sensing or a 6-DOF tracker, we do not observe the state of objects within the gripper.

\autoref{tab:method-observation} describes the observation spaces, \autoref{tab:method-reward} describes the reward formulations, and \autoref{tab:method-task-success} describes the task success criteria (i.e., the condition for a terminal reward) for policy training in simulation. In addition,
\autoref{tab:sim-randomization} describes the randomization range used for training.

\subsubsection{Simulation-Aware Policy Update}

\begin{figure}[H]
    \centering
    \includegraphics[width=0.45\textwidth]{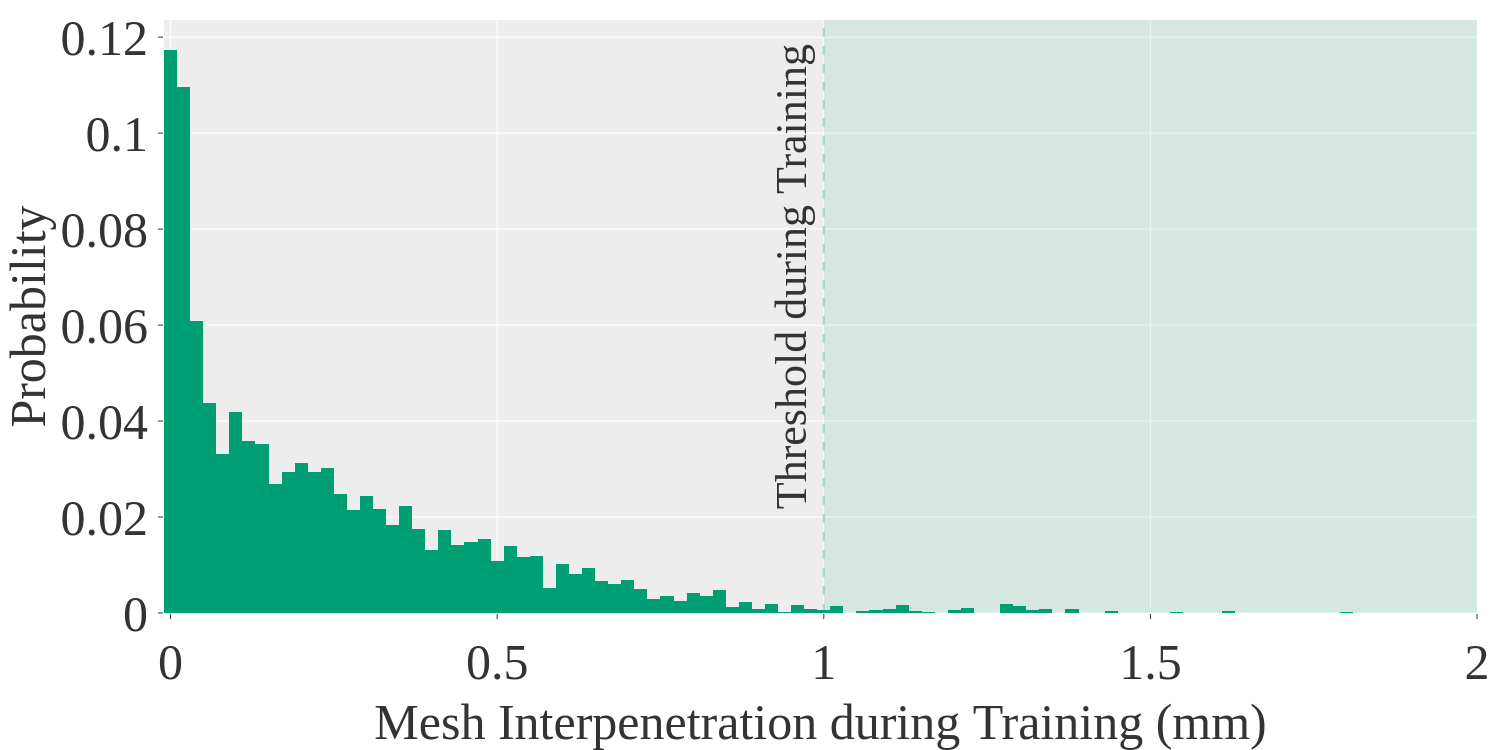}
    \caption{Histogram of mesh interpenetrations during typical training episode.}
    \label{fig:sim_ip_checking}
\end{figure}

\begin{figure}[H]
\centering      
\includegraphics[width=.45\textwidth]{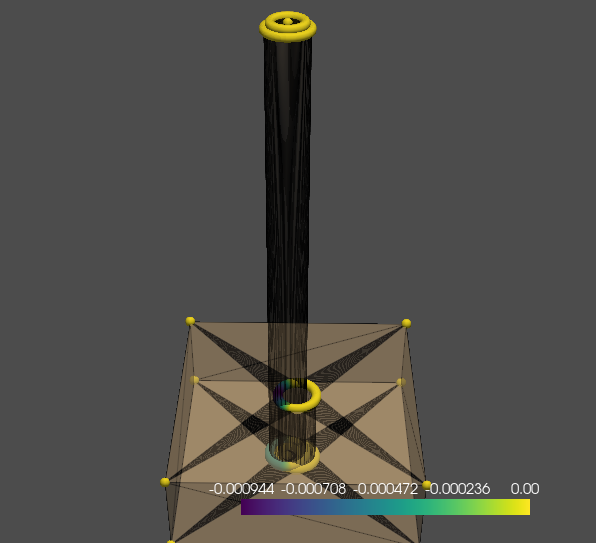}
\caption{Visualization of a transient interpenetration event between peg and hole assets with low-quality meshes. Yellow spheres denote mesh vertices. Colorbar is in m.}
\label{fig:sim-ip-example}
\end{figure}

\autoref{fig:sim_ip_checking} shows the distribution of mesh interpenetration distances during a typical training episode. 
\autoref{fig:sim-ip-example} shows an example of mesh interpenetration between a peg and a hole.

\subsubsection{Joint Evaluation in Simulation}

\autoref{tab:sim_eval_peg_gear} evaluates how the \textbf{Insert} policies perform under different randomization and observation noise conditions in simulation.

\subsubsection{Additional Real Parameters \& Results}
\autoref{tab:latency} describes physics and control frequencies for simulation and the real world, \autoref{tab:reach_eval_params} describes real-world deployment parameters for the \textbf{Reach} policy, and \autoref{tab:eval-real} describes real-world deployment parameters for the remaining policies.

\begin{table}[h]
\centering
\scalebox{.95}{\rowcolors{2}{white}{Gainsboro} 
\begin{tabular}{ lcc } 
\toprule
Task & Success Criterion & Condition \\ \midrule
Pick & Successful lift & $h_{\mathrm{obj}}>3*h_{\mathrm{obj}}+h_{\mathrm{table}}$ \\
Place & Close placement & $||\mathbf{k}^{\mathrm{curr}}-\mathbf{k}^{\mathrm{goal}}||^2<\epsilon_k$ \\
Insert & Peg inserted in hole & $\Delta h<\epsilon_h $ \& $||\mathbf{k}^{\mathrm{plug}}-\mathbf{k}^{\mathrm{socket}}||^2<\epsilon_k$ \\
Gear & Gear inserted on shaft & $\Delta h<\epsilon_h $ \& $||\mathbf{k}^{\mathrm{gear}}-\mathbf{k}^{\mathrm{shaft}}||^2<\epsilon_k$\\
\bottomrule
\end{tabular}}
\caption{Success criterion for each policy. Symbols $h_{obj}$ and $h_{table}$ denote heights of the object and table, $\Delta h$ denotes height from the hole/shaft base to the peg/gear base, $\mathbf{k}$ denotes keypoint positions, $\epsilon_k=10~cm$ is the keypoint distance error threshold, and $\epsilon_h=3~mm$ is the height error threshold.}
\label{tab:method-task-success}
\end{table}

\begin{table*}[h]
\centering
\rowcolors{3}{white}{Gainsboro} 
\begin{tabular}{ llcccccccc } 
\toprule
\multirow{2}{*}{Input} & \multirow{2}{*}{Dimensionality} &\multicolumn{2}{c}{\textbf{Pick}} & \multicolumn{2}{c}{\textbf{Place}} & \multicolumn{2}{c}{\textbf{Insert (Pegs)}} & \multicolumn{2}{c}{\textbf{Insert (Gears)}}\\ 
& & Actor & Critic & Actor & Critic & Actor & Critic & Actor & Critic \\ \midrule
Arm joint angles & 7 & \cmark & \cmark & \cmark & \cmark & \cmark & \cmark & \cmark & \cmark \\
Fingertip pose & 3 (position) + 4 (quaternion) & \cmark & \cmark & \cmark & \cmark & \cmark & \cmark & \cmark & \cmark \\
Object (peg/gear) pose & 3 (position) + 4 (quaternion) & \cmark & \cmark & & \cmark & & \cmark & & \cmark \\
Target pose & 3 (position) + 4 (quaternion) & & & \cmark & \cmark & & \cmark & & \cmark \\
Target pose with noise & 3 (position) + 4 (quaternion) & & & & & \cmark & & \cmark & \\
Relative target pose with noise & 3 (position) + 4 (quaternion) & & & & & \cmark & & \cmark & \\
\hline
Arm joint velocities & 7 & & \cmark &  & \cmark &  & \cmark &  & \cmark \\
Fingertip linear velocity & 3 & & \cmark & & \cmark & & \cmark & & \cmark \\
Fingertip angular velocity & 3 & & \cmark & & \cmark & & \cmark & & \cmark \\
Object (peg/gear) linear velocity & 3 & & & & \cmark & & \cmark & & \cmark \\
Object (peg/gear) angular velocity & 3 & & & & \cmark & & \cmark & & \cmark \\
Target position observation noise & 3 & & & & & & \cmark & & \cmark \\
Relative target pose & 3 (position) + 4 (quaternion) & & & & & & \cmark & & \cmark \\
\bottomrule

\end{tabular}

\caption{Observations provided to the actor and critic for each policy. For \textbf{Pick}, \textit{target pose} is the current plug pose; for \textbf{Place}, \textit{target pose} is the target plug pose; and for \textbf{Insert (Pegs/Gears)}, \textit{target pose} is a target end-effector pose.} 
\label{tab:method-observation}
\end{table*}

\begin{table*}[h]
\centering
\begin{tabular}{ lcccccccc } 
\toprule
\rowcolor{Gainsboro}Per-Timestep Reward & Formulation & Scale & Justification & \textbf{Reach} & \textbf{Pick} & \textbf{Place} & \textbf{Insert (Pegs)} &\textbf{Insert (Gears)}\\ \midrule
Distance to goal &$-||\mathbf{k}^{\mathrm{curr}}-\mathbf{k}^{\mathrm{goal}}||^2$  & 1.0 & Move closer to goal & \cmark & \cmark & \cmark &  &  \\
SDF-based distance &$-\log(\Sigma_{i}^{N}\phi(x_i)/N)$  & 10.0 & Align shapes & & &  & \cmark & \cmark \\
\midrule
\rowcolor{Gainsboro}Scaling Factor on Return & & Range & & & & & & \\ \midrule
Mesh interpenetration & $1-\tanh(d/\epsilon_{d})$ & $\left[0, 1\right]$ & Avoid interpenetration & & & & \cmark & \cmark \\
Randomization range & $\zeta_{\mathrm{curr}}/\zeta_{\mathrm{max}}$ & $\left[0,1\right]$ & Adapt to randomization & & & & \cmark & \cmark \\
Curriculum difficulty & $\emph{D}_{\mathrm{curr}}/\emph{D}_{\mathrm{max}}$ & $\left[0,1\right]$ & Adapt to difficulty & & & & \cmark & \cmark \\
Bonus scale & $1/(\bigtriangleup \emph{h}+0.1)$ & $\left[0,1\right]$ & Move closer to goal & & & & \cmark & \cmark \\
\midrule
\rowcolor{Gainsboro}Sparse Reward & Condition & Value & & & & & & \\ \midrule
Task success & See \autoref{tab:method-task-success} & 10.0 & Complete task & & \cmark & \cmark & \cmark & \cmark \\
Close to target & $||\mathbf{k}^{\mathrm{curr}}-\mathbf{k}^{\mathrm{goal}}||^2<\epsilon_k$ & 1.0 & Move close to target & & \cmark & \cmark &  &  \\
\bottomrule
\end{tabular}
\caption{Rewards for each policy. Symbol $\mathbf{k}$ denotes keypoint positions, $x_i$ denotes vertices on the plug, $N$ is the number of vertices sampled from the plug mesh, $\phi(x_i)$ is the SDF value at a given vertex, $d$ is the mean mesh interpenetration distance, $\epsilon_d$ is the interpenetration distance threshold, $\zeta$ is the magnitude of the randomization range, $D$ is the stage of the curriculum, $\Delta h$ is the height difference between current pose and goal pose when the episode is marked as successful, and $\epsilon_k$ is the keypoint distance error threshold. Note that the reward formulations for the \textbf{Reach}, \textbf{Pick}, and \textbf{Place} policies are simple; those for \textbf{Insert (Pegs)} and \textbf{Insert (Gears)} required more nuance.}
\label{tab:method-reward}
\end{table*}

\begin{table*}[h]
\rowcolors{3}{white}{Gainsboro} 
\centering
\begin{tabular}{ ll|ll|ll|ll } 
\toprule
\multicolumn{2}{c}{\textbf{Reach}} & \multicolumn{2}{c}{\textbf{Pick} \& \textbf{Place}} & \multicolumn{2}{c}{\textbf{Insert (Pegs)}} & \multicolumn{2}{c}{\textbf{Insert (Gears)}}\\ \midrule
Parameter & Range & Parameter & Range & Parameter & Range & Parameter & Range \\\midrule
Hand X-axis & [0.3, 0.7] m & Hand X-axis & [0.3, 0.7] m & Plug to socket $\Delta XY$ & [-10, 10] mm & Gear to shaft $\Delta XY$ & [-10, 10] mm \\
Hand Y-axis & [-0.35, 0.35] m & Hand Y-axis & [-0.35, 0.35] m & Plug to socket $\Delta Z$ & [0, 20] mm & Gear to shaft $\Delta Z$ & [0, 20] mm \\
Hand Z-axis & [0.05, 0.35] m & Hand Z-axis & [0.05, 0.35] m & Socket X-axis & [0.4, 0.6] m & Shaft X-axis & [0.4, 0.6] m  \\
Hand rotation & [-15, 15] deg & Hand rotation & [-15, 15] deg & Socket Y-axis & [-0.1, 0.1] m & Shaft Y-axis & [-0.1, 0.1] m  \\
Target X-axis & [0.35, 0.65] m & Object X-axis & [0.35, 0.65] m & Socket Z-axis & [0.0, 0.05] m & Shaft Z-axis & [0.0, 0.05] m  \\
Target Y-axis & [-0.15, 0.15] m & Object Y-axis & [-0.15, 0.15] m & Socket yaw & [-5, 5] deg & Shaft yaw & [-15, 15] deg  \\
Target Z-axis & [0.0, 0.05] m & Object Z-axis & [0.0, 0.05] m & Observation noise & [-1, 1] mm & Observation noise & [-1, 1] mm \\\bottomrule
\end{tabular}
\caption{Ranges for initial pose randomization and observation noise during training. Values were uniformly sampled.}
\label{tab:sim-randomization}
\end{table*}

\begin{table*}[t]
    \begin{minipage}{.5\linewidth}
    \centering
    \scalebox{0.8}{
    \begin{tabular}{@{}lccccc@{}}
    \toprule
    & \multirow{2}{*}{\textbf{Physics}} & \multicolumn{2}{c}{\textbf{Observations}} & \multicolumn{2}{c}{\textbf{Actions}}\\
    & & Low-Level & Policy & Low-Level & Policy  \\\midrule
    \textbf{Simulation}  & 120 Hz & 60 Hz & 60 Hz & 60 Hz  &  60 Hz \\\midrule
    \textbf{Real-world} & $c/L$ & 1000 Hz & 60 Hz & 1000 Hz & 60 Hz\\\bottomrule
    \end{tabular}
    }
    \caption{Physics and control frequencies for simulation and reality. Physics frequency in the real world is given by $c/L$, where $c$ is the speed of sound and $L$ is a characteristic length-scale; $c$ can be approximated by the Newton-Laplace equation.}
    \label{tab:latency}
    \end{minipage}%
    \begin{minipage}{.5\linewidth}
      \centering
\scalebox{0.9}{
\begin{tabular}{ llll } 
\toprule
\textbf{Deploy Method} & \textbf{Parameter} & \textbf{Value} \\ \midrule
All &  Controller Gains & [1000, 1000, 1000, 50, 50, 50]\\
PLAI &  Pos. Action Scale & [0.0005, 0.0005, 0.0005]\\
PLAI &  Rot. Action Scale & [0.001, 0.001, 0.001]\\
Nominal &  Pos. Action Scale & [0.01, 0.01, 0.01]\\
Nominal &  Rot. Action Scale & [0.01, 0.01, 0.01]\\
\bottomrule
\end{tabular}}
\caption{Reach policy evaluation parameters in real robot}
\label{tab:reach_eval_params}
    \end{minipage} 
\end{table*}

\begin{table*}[h]
\centering
\scalebox{0.92}{
\begin{tabular}{ l|cccc|cccc } 
\toprule
\multirow{2}{*}{\textbf{Randomization}} & \multicolumn{4}{c}{\textbf{Insert (Pegs)}} & \multicolumn{4}{c}{\textbf{Insert (Gears)}} \\
 & Success (\%) & Engage (\%) & Pos. Err. (mm) & Rot. Err. (rad) & Success (\%) & Engage (\%) & Pos. Err. (mm) & Rot. Err. (rad)\\\midrule
\rowcolor{gray(x11gray)} Peg/Gear Pos. (mm) & &&&&&&& \\\midrule
$\left[0,0\right]$ &92.40$\pm$2.30 & 98.40$\pm$1.67 & 2.83$\pm$0.46 & 0.075$\pm$0.0058 & 82.40$\pm$3.58 &84.40$\pm$3.78&10.92$\pm$1.16 &0.30$\pm$0.055 \\
\rowcolor{Gainsboro} $\left[-5,5\right]$ &92.40$\pm$1.52&97.80$\pm$1.48 & 2.98$\pm$0.42 & 0.075$\pm$0.0083 &78.40$\pm$3.97 &81.80$\pm$3.27 & 12.32$\pm$2.14& 0.26$\pm$0.042\\
$\left[-10,10\right]$ & 88.60$\pm$2.41&96.60$\pm$0.023 &3.80$\pm$0.00080 & 0.086$\pm$0.026 & 82.00$\pm$3.54 &85.20$\pm$2.68 &11.09$\pm$1.92 &0.29$\pm$0.040 \\
\rowcolor{Gainsboro} $\left[-15,15\right]$ & 75.20$\pm$6.49&89.60$\pm$3.91 &46.16$\pm$38.71 &0.45$\pm$0.25 & 71.88$\pm$2.62&80.68$\pm$1.36 &12.90$\pm$1.16 & 0.22$\pm$0.075\\
$\left[-20,20\right]$ & 69.00$\pm$5.10 &85.4$\pm$5.81 &53.12$\pm$38.92 &0.33$\pm$ 0.059&71.60$\pm$1.33 &80.88$\pm$0.85 &12.66$\pm$1.24 & 0.22$\pm$0.016\\\midrule
\rowcolor{gray(x11gray)} Hole/Shaft Pos. (cm) & &&&&&&& \\\midrule
$\left[0,0\right]$ &85.00$\pm$2.49&95.00$\pm$1.58&3.17$\pm$0.30&0.074$\pm$0.011&78.20$\pm$3.56&79.20$\pm$3.70&19.75$\pm$7.21&0.40$\pm$0.024 \\
\rowcolor{Gainsboro}$\left[-5,5\right]$ &89.80$\pm$0.015&97.40$\pm$0.015&2.58$\pm$0.36&0.065$\pm$0.019&82.60$\pm$4.34&84.60$\pm$3.85&14.22$\pm$2.12& 0.30$\pm$0.16\\
$\left[-10,10\right]$ & 88.60$\pm$2.41&96.60$\pm$0.023 &3.80$\pm$0.00080 & 0.086$\pm$0.026 & 82.00$\pm$3.54 &85.20$\pm$2.68 &11.09$\pm$1.92 &0.29$\pm$0.040 \\
\rowcolor{Gainsboro}$\left[-15,15\right]$ &78.60$\pm$2.07&86.20$\pm$1.92&7.10$\pm$0.44&0.085$\pm$0.0052&67.86$\pm$3.47&75.88$\pm$2.58&14.37$\pm$1.84&0.22$\pm$0.0088 \\
$\left[-20,20\right]$ &69.40$\pm$4.16&78.00$\pm$2.55&10.38$\pm$1.14&0.11$\pm$0.0065&67.88$\pm$2.56&73.18$\pm$2.36&15.59$\pm$1.88&0.25$\pm$0.016 \\\midrule
\rowcolor{gray(x11gray)} Observation Noise (mm) & &&&&&&& \\\midrule
$\left[0,0\right]$ &87.00$\pm$2.55&93.60$\pm$1.52&4.84$\pm$.51&0.091$\pm$0.021&81.80$\pm$1.64&84.00$\pm$0.020&11.94$\pm$1.09&0.28$\pm$0.045 \\
\rowcolor{Gainsboro}$\left[-1,1\right]$ & 88.60$\pm$2.41&96.60$\pm$0.023 &3.80$\pm$0.00080 & 0.086$\pm$0.026 & 82.00$\pm$3.54 &85.20$\pm$2.68 &11.09$\pm$1.92 &0.29$\pm$0.040 \\
$\left[-2,2\right]$ &86.80$\pm$2.28&97.00$\pm$2.35&3.89$\pm$0.56&0.085$\pm$0.011&79.40$\pm$2.70&82.20$\pm$4.09&11.40$\pm$1.49&0.27$\pm$0.044 \\
\bottomrule
\end{tabular}
}
\caption{Joint evaluation of Simulation-Based Policy Update, SDF-Based Reward, and Sampling-Based Curriculum. (A) \textbf{Pegs and Holes} assembly \textbf{Insert} policy. (B) \textbf{Gears and Gearshafts} assembly \textbf{Insert} policy. \textit{Engage} denotes partial insertion. Policies were trained with moderate randomization (i.e., plug randomization of $\pm$ 10 mm, socket  randomization of $\pm$ 10 cm, and observation noise of $\pm$ 1 mm); thus, the table evaluates in-distribution and out-of-distribution performance. Each test was executed on 5 seeds, with 1000 trials each.}
\label{tab:sim_eval_peg_gear}
\end{table*}

\begin{table*}[t]
\rowcolors{2}{white}{Gainsboro} 
\centering

\begin{tabular}{ l|l|l|l|l|l } 
\toprule
& \textbf{Reach} & \textbf{Pick} & \textbf{Place} & \textbf{Insert (Pegs)} & \textbf{Insert (Gears)}\\ \midrule
MLP network size (Actor) & [256, 128, 64] & [256, 128, 64] & [256, 128, 64] & [512, 256, 128] & [512, 256, 128] \\
MLP network size (Critic) & [512, 256, 128] & [512, 256, 128] & [512, 256, 128] & [256, 128, 64] & [256, 128, 64] \\
LSTM network size (Actor) & - & - & - & 256 & 256 \\
Horizon length (T)       & 128 & 128 & 128 & 256 & 256 \\
Adam learning rate       & 1e-4 & 1e-4 & 1e-4 & 1e-3 & 1e-3 \\
Discount factor ($\gamma$)& 0.99 & 0.99 & 0.99 & 0.998 & 0.998 \\
GAE parameter ($\lambda$) & 0.95 & 0.95 & 0.95 & 0.95 & 0.95 \\
Entropy coefficient      & 0.0 & 0.0 & 0.0 & 0.0 & 0.0 \\
Critic coefficient       & 2 & 2 & 2 & 2 & 2 \\
Minibatch size           & 2048 & 2048 & 2048 & 2048 & 2048 \\
Minibatch epochs       & 8 & 8 & 8 & 8 & 8 \\
Clipping parameter ($\epsilon$) & 0.2 & 0.2 & 0.2 & 0.2 & 0.2 \\\bottomrule
\end{tabular}
\caption{PPO parameters and network architectures used in each policy.}
\label{tab:ppo-param}
\end{table*}

\begin{table*}[t]
\rowcolors{2}{white}{Gainsboro}
\centering
\begin{tabular}{ llll } 
\toprule
\textbf{Task} & \textbf{Asset} & \textbf{Parameter} & \textbf{Value} \\ \midrule
Pick \& Place & All & Controller Gains & [1000, 1000, 1000, 50, 50, 50]\\
Pick \& Place & All & Pos. Action Scale & [0.002, 0.002, 0.0015]\\
Pick \& Place & All & Rot. Action Scale & [0.004, 0.004, 0.004]\\\midrule
Insert & Pegs \& Connectors & Controller Gains & [1000, 1000, 100, 50, 50, 50]\\
Insert & Pegs & Pos. Action Scale & [0.0006, 0.0006, 0.0004]\\
Insert & Connectors & Pos. Action Scale & [0.0004, 0.0004, 0.0004] \\
Insert & Pegs \& Connectors & Rot. Action Scale & [0.001, 0.001, 0.001]\\
Insert & Pegs & Leaky PLAI Threshold &  [0.05, 0.05, 0.03]\\
Insert & Connectors & Leaky PLAI Threshold & [0.04, 0.04, 0.05]\\\midrule
Insert & Gears & Controller Gains & [300, 300, 300, 50, 50, 50]\\
Insert & Gears & Pos. Action Scale & [0.0005, 0.0005, 0.0004]\\
Insert & Gears & Rot. Action Scale & [0.001, 0.001, 0.001]\\
Insert & Gears & Leaky PLAI Threshold & [0.05, 0.05, 0.05]\\\midrule
Insert & All & Observation Noise & XY-axes: [-2, 2]~mm\\
Insert & All & Pos. Randomization &  XY-axes: [-10, 10]~mm\\
Insert & All & Height Randomization &  [10, 20]~mm above socket/shaft\\
Insert & All & Rot. Randomization &  Yaw: [-5, 5]~deg\\
Insert & All & Clearance &  [0.5, 0.6]~mm\\\bottomrule
\end{tabular}

\caption{Real-world evaluation parameters for \textbf{Insert}. \textit{Action scales} are scalars applied to position and rotation actions in order to bring robot execution speeds to within comfortable limits.}
\label{tab:exp-params-real}
\end{table*}

\end{document}